\documentclass[sigconf]{acmart}

\usepackage{subfigure,multirow}
\usepackage{graphicx}
\usepackage{xurl}
\usepackage{color}

\usepackage{array}
\usepackage{amsmath}
\usepackage{mathrsfs} 
\usepackage{autobreak}
\usepackage{algorithm}
\usepackage[noend]{algorithmic}
\usepackage{dsfont}
\renewcommand{\algorithmiccomment}[1]{//#1}

\begin{document}

\acmYear{2022}\copyrightyear{2022}
\setcopyright{acmlicensed}
\acmConference[ACM MobiCom '22]{The 28th Annual International Conference On Mobile Computing And Networking}{October 17--21, 2022}{Sydney, NSW, Australia}
\acmBooktitle{The 28th Annual International Conference On Mobile Computing And Networking (ACM MobiCom '22), October 17--21, 2022, Sydney, NSW, Australia}
\acmPrice{15.00}
\acmDOI{10.1145/3495243.3560551}
\acmISBN{978-1-4503-9181-8/22/10}



\title[]{Real-time Neural Network Inference on Extremely Weak Devices: Agile Offloading with Explainable AI}

\author{Kai Huang}
\affiliation{%
	\institution{University of Pittsburgh}
	\country{USA}
}
\email{k.huang@pitt.edu}

\author{Wei Gao}
\affiliation{%
	\institution{University of Pittsburgh}
	\country{USA}
}
\email{weigao@pitt.edu}

\begin{abstract}	
	With the wide adoption of AI applications, there is a pressing need of enabling real-time neural network (NN) inference on small embedded devices, but deploying NNs and achieving high performance of NN inference on these small devices is challenging due to their extremely weak capabilities. Although NN partitioning and offloading can contribute to such deployment, they are incapable of minimizing the local costs at embedded devices. Instead, we suggest to address this challenge via agile NN offloading, which migrates the required computations in NN offloading from online inference to offline learning. In this paper, we present \emph{AgileNN}, a new NN offloading technique that achieves real-time NN inference on weak embedded devices by leveraging eXplainable AI techniques, so as to explicitly enforce feature sparsity during the training phase and minimize the online computation and communication costs. Experiment results show that AgileNN's inference latency is $>$6$\times$ lower than the existing schemes, ensuring that sensory data on embedded devices can be timely consumed. It also reduces the local device's resource consumption by $>$8$\times$, without impairing the inference accuracy.
\end{abstract}
\maketitle


\section{Introduction}
Neural networks (NNs) have been used to enable many new applications, such as face and speech recognition \cite{hu2015face, amodei2016deep}, object tracking \cite{ciaparrone2020deep,ahmed2015improved}, and personal assistants for business \cite{yu2019deep} and health \cite{zhao2021deep}. With the penetration of these applications into our daily life, there is a pressing need of enabling real-time NN inference on small embedded devices, to allow more intelligent and prompt decision making on these weak devices. For example, on-device data processing on home security sensors \cite{zhao2008low} and industry actuators \cite{progressiveautomations} will allow prompt response to sporadic events, and real-time analysis of human activity data on wearables could timely identify potential health risks \cite{bui2019ebp,abdelhamid2020self}. Deployment of NN models on small drones and robots is the technical foundation of these devices' autonomous navigation \cite{gao2017intention, ayyalasomayajula2020deep}, which is useful in many environment surveillance, disaster rescue and military scenarios. Furthermore, real-time NN inference, if made possible on energy-harvesting-powered sensors \cite{iyer2016inter,gobieski2019intelligence} and RF-powered devices \cite{kellogg2014wi,ma2018enabling}, could expand the current horizon of AI to another magnitude.

Deploying NNs on these small devices, however, is very challenging due to the disparity between these devices' weak capabilities and NNs' high computing demands. For example, the ResNet50 model contains 23 million parameters and 50 convolutional layers \cite{he2016deep}, and requires at least 100MB memory and a processor of $>$2GHz to achieve 60ms inference latency on a smartphone \cite{niu201926ms}. Such amount of computing resources, however, is $>$10 times higher than what is available on a STM32 microcontroller (MCU)\footnote{The STM32 MCUs have been widely used on embedded sensors and actuators. The STM32F746 MCU, for example, is equipped with an ARM Cortex-M7 processor running at 216 MHz and 320KB of local memory \cite{stm32f7}.}.

\begin{figure}
	\centering
	\includegraphics[width=0.95\linewidth]{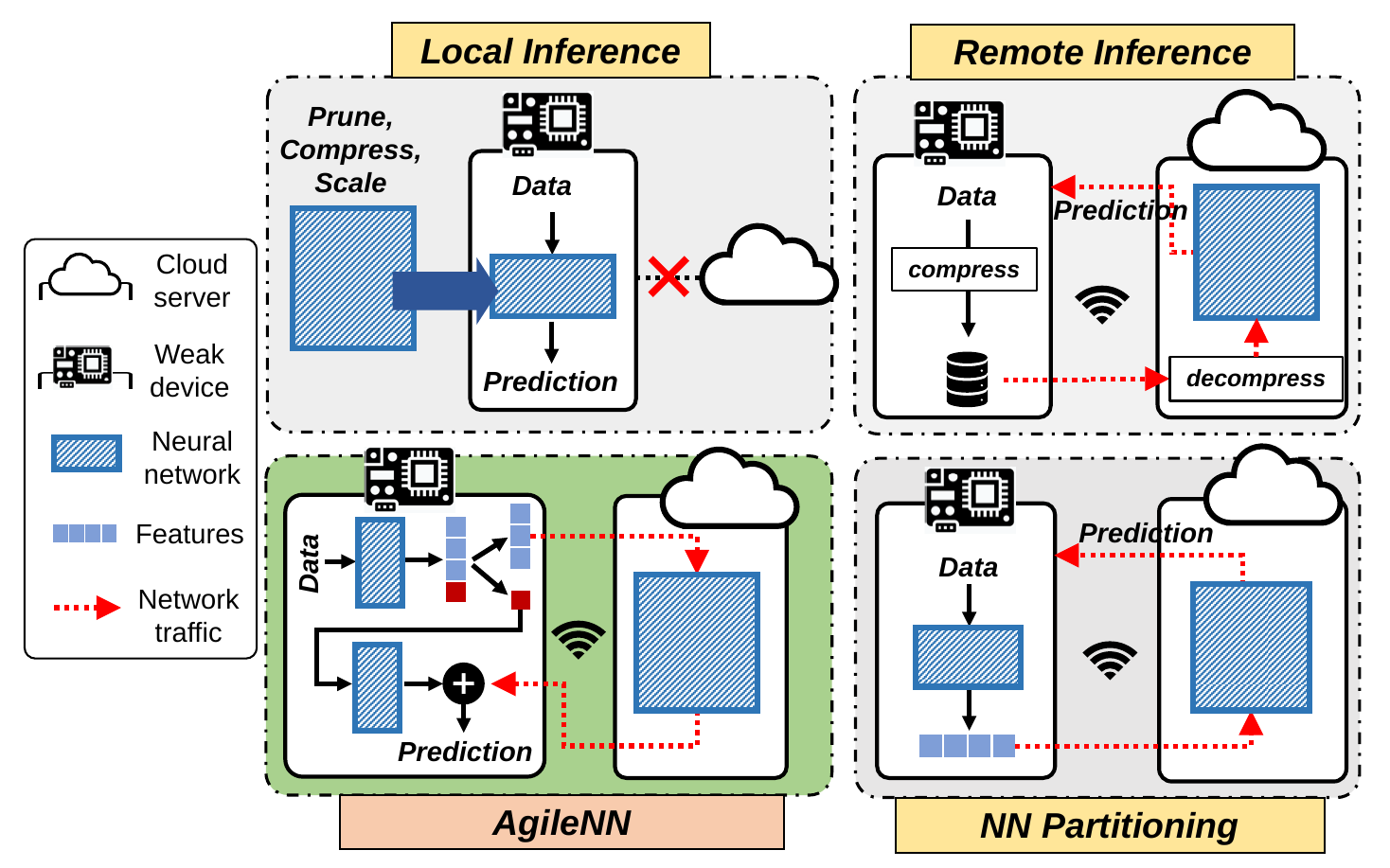}
	\vspace{-0.1in}
	\caption{Existing work vs. AgileNN}
	\vspace{-0.15in}
	\label{fig:comparison}
\end{figure}

\begin{table*}[ht]
	\begin{tabular}{|ll||l|l|l|l|l|}
		\hline
		\multicolumn{2}{|c||}{}                                                & Local comput. & Memory & Data & Inference & Training \\ 
		\multicolumn{2}{|c||}{} & complexity & cost & trans. cost & accu. loss & cost\\
		\hline\hline
		\multicolumn{1}{|c|}{\multirow{3}{*}{Local Inference}}  & Compression \cite{denton2014exploiting, gong2014compressing} & Very High          & High              & None               & High  & High         \\ \cline{2-7} 
		\multicolumn{1}{|c|}{}                                  & Pruning \cite{niu2020patdnn, han2015learning}    & Very High          & High              & None               & High  & High        \\ \cline{2-7} 
		\multicolumn{1}{|c|}{}                                  & NAS \cite{lin2020mcunet, banbury2021micronets}        & High        & Medium            & None               & Medium   & Very High       \\ \hline
		\multicolumn{1}{|c|}{\multirow{2}{*}{Remote Inference}} & JPEG \cite{wallace1992jpeg}, MPEG \cite{ le1991mpeg}  & Low           & Low               & High              & Low  & Low         \\ \cline{2-7} 
		\multicolumn{1}{|c|}{}                                  & NN-favorable compression \cite{liu2018deepn,liu2019machine} & Medium           & Low               & Medium            & Medium     & Medium      \\ \hline
		\multicolumn{2}{|c||}{NN Partitioning \cite{kang2017neurosurgeon,  hu2019dynamic, li2018jalad, yao2020deep, laskaridis2020spinn,ko2018edge}}                                 & High          & Medium              & Low          & Low  & Medium         \\ \hline\hline
		\multicolumn{2}{|c||}{\textbf{AgileNN}}                                & \textbf{Very Low}  & \textbf{Low}      & \textbf{Very Low} & \textbf{Very Low} & \textbf{Medium} \\ \hline
	\end{tabular}
	\vspace{0.1in}
	\caption{Comparison of the approaches to NN inference on weak devices}
	\vspace{-0.25in}
	\label{table:comparison}
\end{table*}

To eliminate such disparity, researchers aimed to reduce the NN complexity via compression \cite{denton2014exploiting, gong2014compressing} or pruning \cite{niu2020patdnn, han2015learning} (Figure \ref{fig:comparison} - top left), which remove redundant NN weights and structures. However, the existing schemes mainly target strong mobile devices (e.g., smartphones) where a moderate reduction of NN complexity is sufficient. When being tailored to the weak embedded devices' extreme resource constraints, the over-simplified NNs will suffer large reductions of inference accuracy. For example, when the size of a ResNet50 model is reduced by 100 times, its inference accuracy could drop from 77\% to 62\% \cite{frankle2020pruning}. Even with the recent Neural Architecture Search (NAS) technique that finds the best NN structure with the given complexity constraint \cite{lin2020mcunet, banbury2021micronets}, the inference accuracy loss could be still $>$10\%.

Instead, a better solution to avoiding the inference accuracy loss is to offload the NN computations to a cloud server. To minimize the communication cost of offloading, one can compress the NN input data \cite{wallace1992jpeg, le1991mpeg, liu2018deepn, liu2019machine} before transmission (Figure \ref{fig:comparison} - top right), but the compression ratio could be limited and result in high data transmission latency, with the low-speed wireless radios (e.g., Bluetooth and ZigBee) used on embedded devices for energy saving purposes. Later research efforts suggest to partition the NN (Figure \ref{fig:comparison} - bottom right), and use the \emph{Local NN}\footnote{In this rest of this paper, we use \emph{Local NN} to indicate the portion of partitioned NN at the local device, and \emph{Remote NN} to indicate the portion of partitioned NN at the cloud server.} to transform the input data into a more compressible form of feature representations before transmission. Existing NN partitioning schemes \cite{kang2017neurosurgeon, hu2019dynamic, li2018jalad, yao2020deep, laskaridis2020spinn,ko2018edge}, however, need to use an expensive Local NN to enforce feature sparsity and incur unacceptable computing latency on the local device. The key reason of this limitation is that these schemes regardlessly apply the same learning approach to every input data, and hence need a sufficient amount of representation power at the local NN for the worst case of input data. 

To address this limitation and practically enable NN inference on extremely weak devices (e.g., MCUs) with the minimum latency, in this paper we present \emph{AgileNN}, a new technique that shifts the rationale of NN partitioning and offloading from fixed to agile and data-centric. Our basic idea is to incorporate the knowledge about different input data's heterogeneity in training, so that the required computations to enforce feature sparsity are migrated from online inference to offline training. More specifically, we interpret such heterogeneity as different data features' importance to NN inference, and leverage the eXplainable AI (XAI) techniques \cite{selvaraju2017grad,sundararajan2017axiomatic} to explicitly evaluate such importance during training. In this way, as shown in Figure \ref{fig:comparison} - bottom left, the online inference can enforce feature sparsity by only compressing and transmitting the less important features, without involving expensive NN computations. The important features, on the other hand, are retained at the local device and can be perceived by a lightweight NN with low complexity. Predictions from Local NN and Remote NN, eventually, are combined at the local device for inference.

The major challenge of using AgileNN in practice, however, is that different data features may have similar importances to NN inference. In this case, sparsity among less important features will be reduced and result in lower data compressibility, and more features also need to be retained at the local device, incurring extra computing latency. To address this challenge and simultaneously minimize the local embedded device's costs in computation and communication, AgileNN's basic approach is to intentionally manipulate the data features' importance via non-linear transformation in the high-dimensional feature space, so as to ensure that such importance's distribution over different features is skewed. In other words, only few features make the majority of contributions to NN inference. In our design, we realize such skewness manipulation with a highly lightweight feature extractor, and jointly train the feature extractor with Local and Remote NNs to ensure inference accuracy.

To our best knowledge, AgileNN is the first technique that achieves real-time NN inference on embedded devices with extremely weak capabilities in computation and communication. Our detailed contributions are as follows:
\begin{itemize}
	\item We effectively migrate the required computations in NN offloading from online inference to offline training, by leveraging XAI techniques that allow lightweight enforcement of feature sparsity at runtime. 
	\item We developed new AI techniques that use XAI to explicitly manipulate the importances of different data features in NN inference, so as to ensure the effectiveness of NN partitioning and offloading.	
	\item By enforcing skewness of such importance's distribution over different features, we allow flexible tradeoffs between the accuracy and cost of NN inference on embedded devices, without incurring any extra computing or storage cost.
\end{itemize}

We implemented AgileNN on a STM32F746 MCU board and a server with an Nvidia RTX A6000 GPU, and evaluated the performance of AgileNN on various popular datasets under different system conditions. From our experiment results, we have the following conclusions: 
\begin{itemize}
	\item AgileNN is \emph{real-time}. Compared to the existing schemes \cite{lin2020mcunet,yao2020deep,laskaridis2020spinn}, AgileNN reduces the NN inference latency by up to 6x, and restrains such latency within 20ms on most datasets. It hence supports real-time NN inference on weak embedded devices, by ensuring that the sensory data can always be timely consumed.
	\item AgileNN is \emph{accurate}. Compared to the existing NN partitioning schemes, AgileNN provides the similar inference accuracy but achieves much higher feature sparsity. Such high sparsity, then, reduces the amount of data transmission in NN offloading by up to 70\%.
	\item AgileNN is \emph{lightweight}. Compared to the current NN inference schemes on embedded devices, AgileNN reduces the local energy consumption by $>$8x, while consuming 1.2x less memory and 5x less storage space.
	\item AgileNN is \emph{adaptive}. It minimizes the performance degradation of NN inference in different embedded device settings and system conditions, even with extremely low computing power and wireless bandwidth.
\end{itemize}

\section{Background \& Motivation}
To help better understand the AgileNN design, we first demonstrate the limitations of the existing NN offloading schemes. Then, we motivate our design by introducing XAI techniques that explicitly evaluate the importance of different features, and highlighting the necessity of having features with skewed importance distributions.


\begin{figure}[ht!]
	\centering
	\vspace{-0.15in}
	\hspace{-0.25in}
	\subfigure[Compressing the raw input data] { 
		\includegraphics[width=0.24\textwidth]{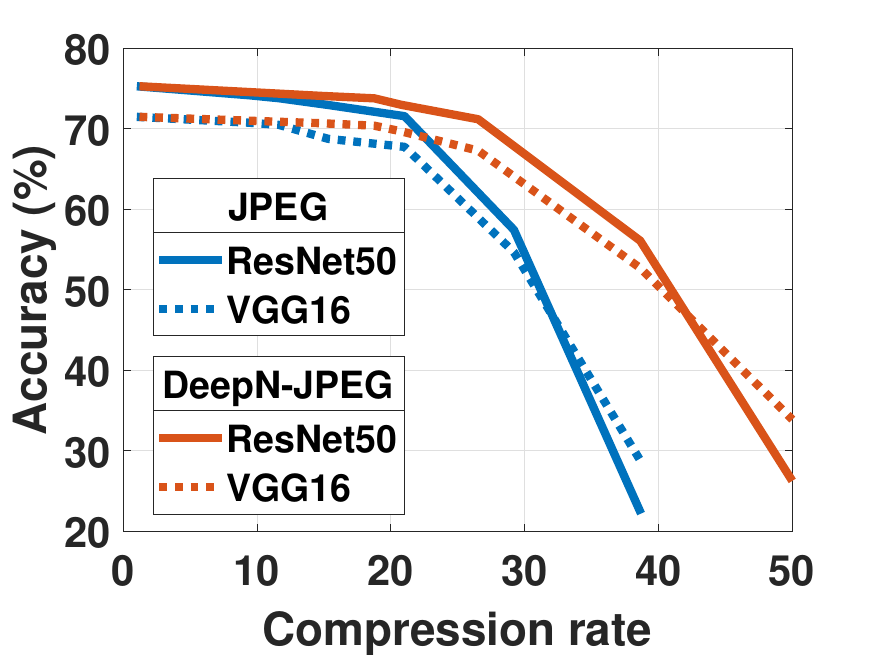}
		\label{fig:compress_raw}
	}
	\subfigure[Feature extraction] { 
		\includegraphics[width=0.24\textwidth]{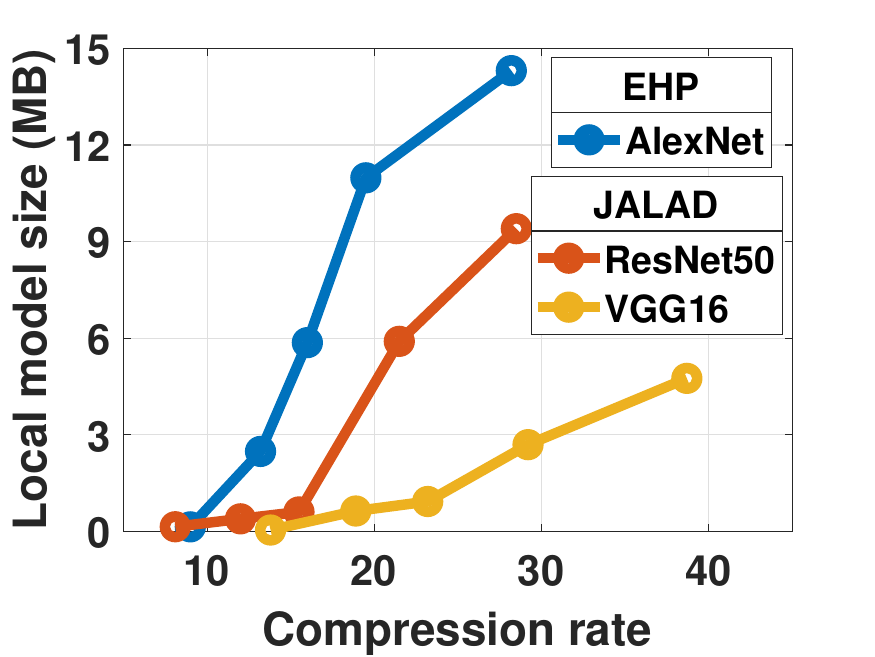}
		\label{fig:cr_model_size}
	}
	\hspace{-0.4in}
	\vspace{-0.15in}
	\caption{Data compressibility in NN offloading}
	\label{fig:compressibility}
	\vspace{-0.1in}
\end{figure}


\subsection{Data Compressibility in NN Offloading}
To reduce the communication cost of NN offloading, an intuitive method is to compress the raw data before transmission, but heavy compression will distort the important information in data and hence affect the NN inference accuracy. To verify such impact, we apply both standard JPEG \cite{wallace1992jpeg} and NN-favorable DeepN-JPEG \cite{liu2018deepn} compression methods to images in the ImageNet dataset \cite{deng2009imagenet}, and measure the inference accuracy loss on various NN models when using different data compression rates \cite{he2016deep,simonyan2014very,sandler2018mobilenetv2}. As shown in Figure \ref{fig:compress_raw}, a moderate compression rate of 25x will reduce the NN inference accuracy by $>$10\%, and such accuracy loss will quickly grow to $>$20\% when the compression rate is $>$30x. 

Instead, current NN partitioning approaches improve the data compressibility by extracting more compressible forms of feature representations from the raw input data. However, as shown in Figure \ref{fig:cr_model_size} with two representative partitioning approaches (EHP \cite{ko2018edge} and JALAD \cite{li2018jalad}), although these schemes can achieve the similar compression rates with the minimum impact on the NN inference accuracy\footnote{The loss of NN inference accuracy in these schemes can be effectively restrained within 1\%, for all the data compression rates being applied.}, their feature extraction is very computationally expensive. For example, achieving a compression rate of 30x will require a large Local NN with a model size of $>$3MB, which is unaffordable on most weak embedded devices such as STM32 MCUs.

\subsection{Explainable AI}
\label{subsec:XAI}


The aforementioned limitation motivates our design that achieves better data compressibility by evaluating different data features' importance during offline training. Based on such knowledge about feature importance, during online inference we can explicitly enforce sparsity in the less important features with the minimum local computing cost. To evaluate such feature importance, classic perturbation-based approaches \cite{breiman2001random} measure how the NN inference accuracy varies after injecting noise to features in all the training data, but cannot precisely evaluate feature importance over individual data samples. Attention-based mechanisms \cite{bahdanau2014neural, vaswani2017attention} support such individualized evaluation by adding an extra weight generator in NN training, but need to tailor the weight generator's structure to each NN model and could hence be inaccurate in some NN models.

\begin{figure}[ht]
	\centering
	\vspace{-0.05in}
	\includegraphics[width=3.4in]{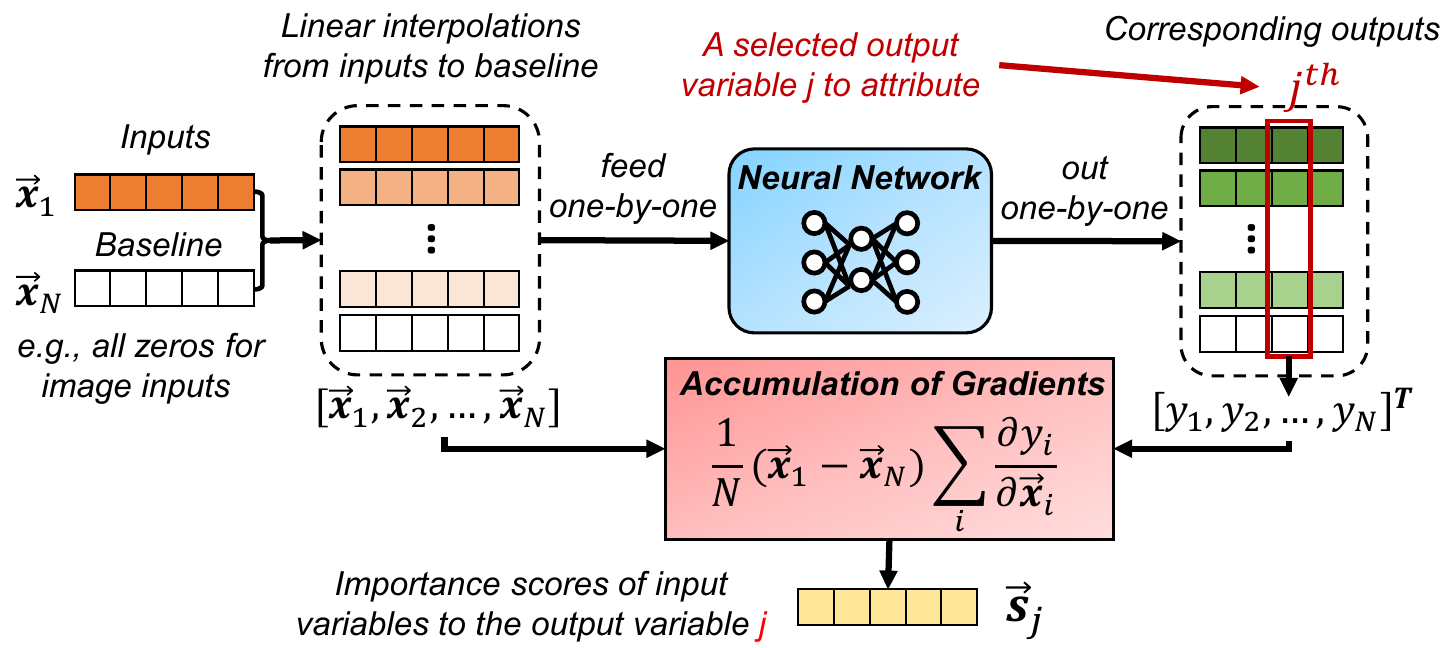}
	\vspace{-0.15in}
	\caption{Integrated Gradients}
	\vspace{-0.1in}
	\label{fig:xai}
\end{figure}



Recent research on eXplainable AI (XAI) improves the accuracy of feature importance evaluation by offering attribution tools that quantitatively correlate each input variable to the NN outputs during training \cite{selvaraju2017grad,sundararajan2017axiomatic}. For example, typical XAI tools such as Integrated Gradients (IG) \cite{sundararajan2017axiomatic}, as shown in Figure \ref{fig:xai}, feed a number of linear interpolations between the input variables and a naive baseline to the NN. Then, for an input variable, they compute each of its interpolation's gradient with respect to the NN's output (e.g., confidence scores), and accumulate these gradients to measure the importance of this input variable\footnote{In practice, such accumulation is used to approximate to the path integral of gradients. The more interpolations are used, the better approximation will be.}. In this way, these XAI tools are robust and applicable to any AI model without extra modification.



\begin{figure}[ht]
	\centering
	\vspace{-0.1in}
	\hspace{-0.25in} 
	\subfigure[Different levels of skewness] { 
		\includegraphics[width=0.24\textwidth]{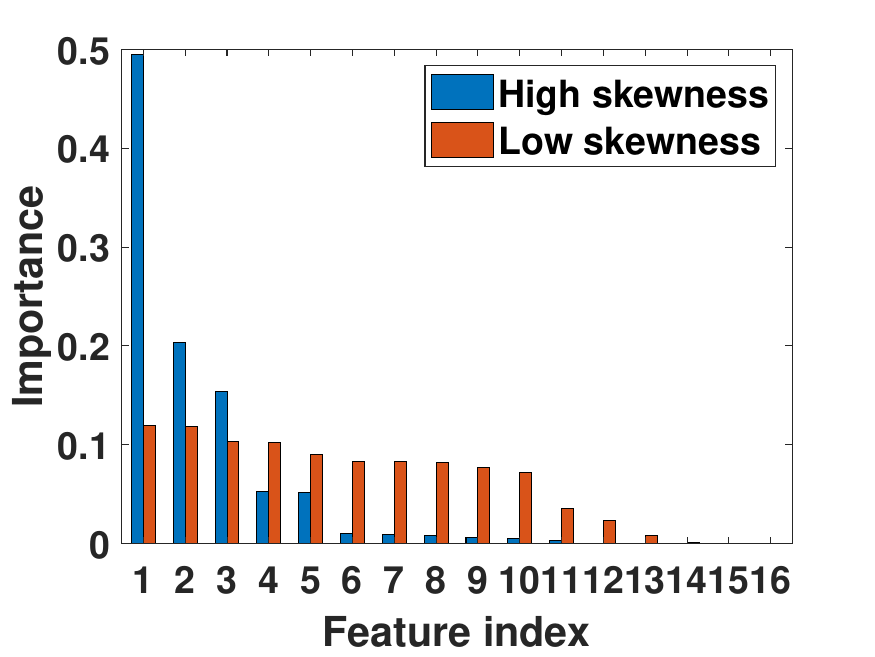}
		\label{fig:different_skewness}
	}
	\subfigure[Skewness CDF] { 
		\includegraphics[width=0.24\textwidth]{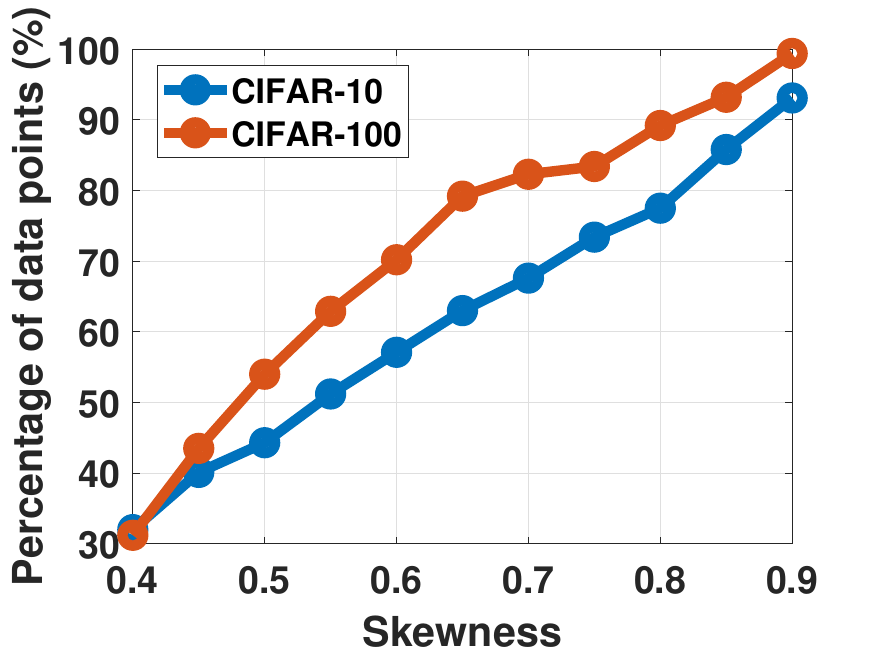}
		\label{fig:original_importance_occupation}
	}
	\hspace{-0.4in}
	\vspace{-0.15in}
	\caption{Skewness of feature importance. Skewness is measured as the normalized importance of the top 20\% features, using the MobileNetV2 model \cite{sandler2018mobilenetv2}.}
	\label{fig:data_characteristics}
	\vspace{-0.1in}
\end{figure}

\begin{figure*}[ht]
	\centering
	\includegraphics[width=7in]{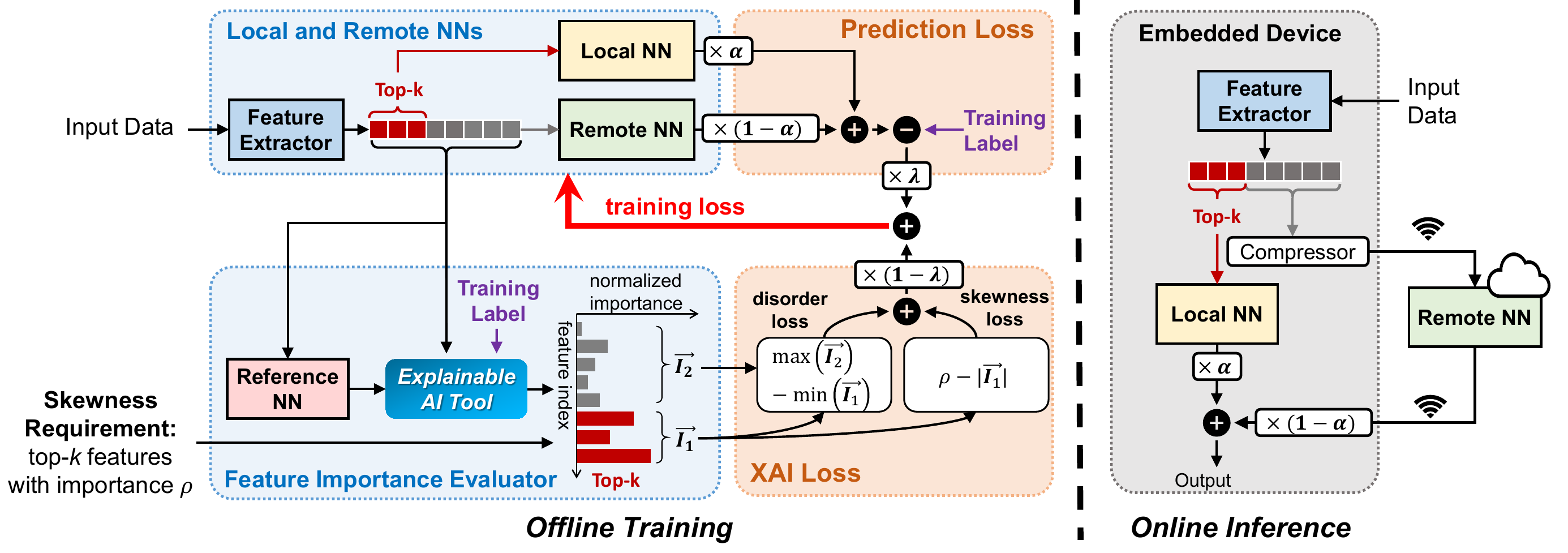}
	\vspace{-0.2in}
	\caption{Overview of AgileNN design}
	\vspace{-0.1in}
	\label{fig:system_overview}
\end{figure*}

However, one key limitation of the existing XAI tools is that its accuracy of feature importance evaluation builds on accurate NN inference in advance. If the NN's output is ambiguous (e.g., due to inadequate training), XAI could produce misleading evaluations because the gradients computed from the NN's output are highly random. In the worst case, such randomness can cause all the features to be misranked by their importance \cite{ribeiro2016model}. This limitation motivates us to use a pre-trained reference NN model in AgileNN's training, to ensure correct XAI evaluation on feature importance.

\vspace{-0.05in}
\subsection{Skewness of Feature Importance}
Based on the feature importance evaluated by XAI, the effectiveness of AgileNN's offloading depends on the skewness of such importance's distribution over different features. The higher such skewness is, the fewer features are playing a dominant role in NN inference and we can hence enforce higher sparsity in less important features without impairing the NN inference accuracy. However, as shown in Figure \ref{fig:different_skewness} that exemplifies such importance distribution of different data samples in the CIFAR-10 dataset \cite{krizhevsky2009learning}, skewness may not always exist in every input data. Furthermore, as shown in Figure \ref{fig:original_importance_occupation}, when we measure skewness as the ratio of normalized importance of the top 20\% features, such skewness in $>$40\% of data samples in the CIFAR-10 and CIFAR-100 datasets \cite{krizhevsky2009learning} is $<$50\%. 

Such low skewness in the input data motivates us to design new NN structures that intentionally manipulate and enhance such skewness in feature extraction, while minimizing the impact of such skewness manipulation on the NN inference accuracy.




\vspace{-0.05in}
\section{System Overview}
As shown in Figure \ref{fig:system_overview}, AgileNN partitions the neural network into a Local NN and a Remote NN. In online inference, AgileNN runtime uses a lightweight feature extractor at the local embedded device to provide feature inputs: the top-$k$ features with high importance are retained by the Local NN to make a local prediction, which is then combined with the Remote NN's prediction from other less important features for the final inference output. In this way, the complexity of Local NN could be minimized without impairing the inference accuracy, and high sparsity can be enforced when compressing and transmitting less important features to the server. 

In offline training, AgileNN jointly trains the feature extractor, Local NN and Remote NN with a unified loss function. In particular, the feature extractor is trained to meet the user's \emph{requirement of feature importance skewness}, such that the normalized importance of top-$k$ features should exceed a threshold $\rho\in[0,1]$. During the training process, enforcing this requirement is equivalent to apply non-linear transformations to the output feature vector in the high-dimensional feature space. 

Based on this design, AgileNN can flexibly balance between the accuracy and cost of NN inference by adjusting the required feature importance skewness. The higher the skewness is (i.e., smaller $k$ and larger $\rho$), the lower resource consumption will be at the local device due to the higher compressibility of less important features being transmitted, but the NN inference is more affected due to the feature extractor's non-linear transformation in the feature space. In practice, with the same AgileNN runtime being trained for the specific embedded device, the user can adaptively choose different tradeoffs according to the application scenarios and local resource conditions, without spending extra local computing or storage resources to maintain multiple NN models \cite{fang2018nestdnn} or adopt different learning strategies \cite{laskaridis2020spinn}.


\subsection{Skewness Manipulation}
In order to manipulate the importance of extracted features and meet the skewness requirement, AgileNN's basic approach is to incorporate both the inference accuracy and current skewness of feature importance into the unified loss function in training. More specifically, in each training epoch, AgileNN feeds the current set of features extracted by the feature extractor to the XAI tool module, which evaluates and outputs the importance of each feature to NN inference. The skewness of feature importance, then, is incorporated into the loss function in the following two aspects.

1) The \emph{disorder loss}, which mandates that the top-$k$ features with highest importance are always in the first $k$ channels of the output feature vector. It is calculated as 
\begin{equation}
L_{\text{disorder}} = \max\left(0, \max(\vec{I}_2) - \min(\vec{I}_1) \right),
\label{eq:disorder_loss}
\end{equation}
where $\vec{I}_1$ indicates the normalized importances of features in the first $k$ channels of the output feature vector and $\vec{I}_2$ indicates the normalized importances of other features. 

This ordering is essential to online inference, where the XAI tool is unavailable and the top-$k$ features with high importance should hence be always located in fixed channels of the extracted feature vector. With such feature ordering, we can further instruct the feature extractor to enhance the importance of the top-$k$ features and hence enforce the required skewness. The knowledge about the fixed locations of top-$k$ features in the feature vector, on the other hand, will also enable prompt split of features for local and remote inferences, without involving any extra computations or manual efforts at run-time. More details of such feature ordering are provided in Section \ref{subsec:feature_ordering}.

2) The \emph{skewness loss}, which measures the difference between the current skewness of feature importance and the skewness requirement. It is calculated as 
\begin{equation}
L_{\text{skewness}}=\max\left(0, \rho - |\vec{I_1}| \right),
\label{eq:skewness_loss}
\end{equation}
where $|\cdot|$ indicates the vector's 1-norm. 

These two loss components are then combined with the standard prediction loss in AgileNN's training. Details about such combined training loss are provided in Section \ref{subsec:combined_loss}.

On the other hand, as described in Section \ref{subsec:XAI}, the accuracy of XAI's feature importance evaluation requires a well-trained NN in advance to provide correct inference labels. Hence, to ensure the quality of AgileNN's training, we introduce a reference NN model, which has been pre-trained for the same learning task with sufficient representation power\footnote{In practice, such reference models are widely available as public online. For example, EfficientNet model is available online \cite{tan2021efficientnetv2} and can achieve $>$90\% inference accuracy on large datasets such as ImageNet \cite{deng2009imagenet}.}, to provide inference outputs to the XAI tool using the extracted features from AgileNN's feature extractor. To further avoid any possible ambiguity, we compare each inference output made by the reference NN with the training label, and only use it in XAI evaluation if the reference NN makes correct predictions.



\begin{figure}[ht]
	\centering
	\vspace{-0.1in}
	\includegraphics[width=0.9\linewidth]{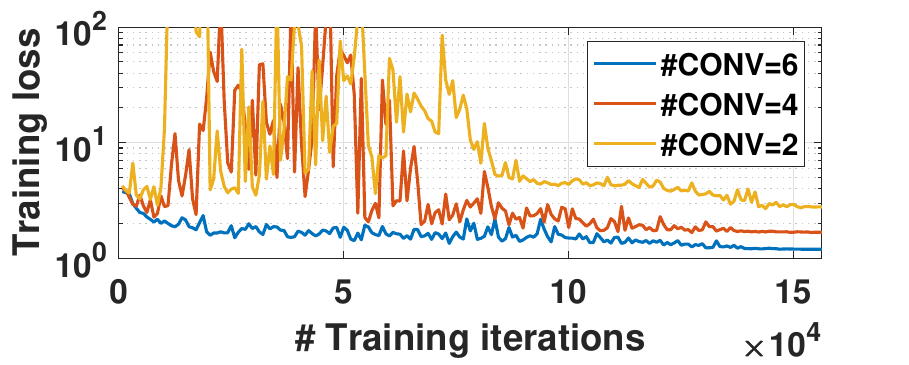}
	\vspace{-0.1in}
	\caption{Training stability with different numbers of convolutional layers in feature extractor}
	\vspace{-0.1in}
	\label{fig:unstable_loss}
\end{figure}

\subsection{Pre-processing the Feature Extractor}
AgileNN jointly trains the Local NN and Remote NN with the feature extractor, so as to ensure that they can provide accurate predictions from the extracted features with skewed distribution of importance. However, since the feature extractor in AgileNN needs to be deployed at the local device and hence has to be very lightweight, it may not have sufficient representation power to meet this learning objective in the initial phase of training, and the joint training may hence encounter unexpectedly high learning difficulty or even fail to converge. For example, as shown in Figure \ref{fig:unstable_loss}, such joint training on CIFAR-100 dataset, if starting from scatch, is highly unstable unless a sufficient number of convolutional layers ($\geq$6) is used in the feature extractor.

To avoid such learning difficulty, AgileNN's approach is to pre-process the feature extractor and initialize its network weights, prior to the joint training with Local and Remote NNs. In this way, the joint training will not start from scratch but instead from a more established stage with less ambiguity, and hence has lower requirement on the initial representation power of the feature extractor.

\begin{figure}[h]
	\centering
	\vspace{-0.1in}
	\includegraphics[width=0.8\linewidth]{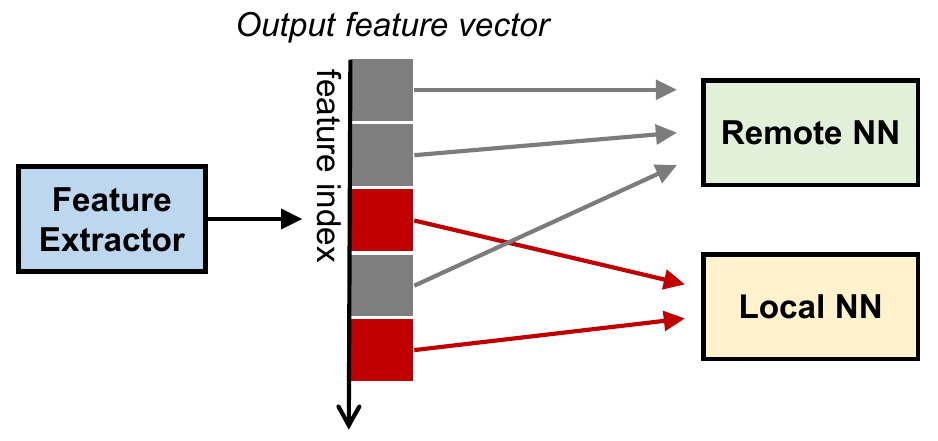}
	\vspace{-0.15in}
	\caption{Pre-processing the feature extractor}
	\vspace{-0.1in}
	\label{fig:connection_initialization}
\end{figure}

More specifically, the feature ordering mandated by the disorder loss in Section 3.1 may be hard to fulfill by the feature extractor in the initial stage of training. Instead, as shown in Figure \ref{fig:connection_initialization}, we select $k$ initial channels in the output feature vector where the top-$k$ features with high importance are most likely to be located. We then use the corresponding $k$ features as the input to the Local NN. More details of selecting these initial channels and integrating such pre-processing into the training process are in Section 5.



\subsection{Combining Local and Remote Predictions}
AgileNN combines the predictions made by the Local and Remote NNs via weighted summation, to produce the final inference output at the local embedded device. {Compared to other NN-based alternatives such as adding an extra NN layer for combination, we use this solution because of the following two reasons. First, computing such point-to-point weighted sums is much more lightweight than NN operations and adds negligible computation overhead to the local device. Second, the outputs of Local and Remote NNs always correspond to the same number of aligned feature channels, and the point-to-point summation retains such alignment. Using an NN layer (e.g., a fully-connected or convolutional layer) to combine these two outputs, on the other hand, could possibly entangle them together and break such alignment, hence impairing the final inference accuracy.

The main difficulty of such combination, however, is that the outputs of Local NN and Remote NN may not be in the same scale and may hence result in extra loss in inference accuracy, because some small but important output values in one NN could be overwhelmed by large values in another NN. To address this difficulty, our solution is to incorporate the summation weight $\alpha$ into the joint training procedure. Being the same as other NN parameters, $\alpha$ is also trained with gradient-based feedback using stochastic gradient descent (SGD) algorithms \cite{bottou2012stochastic}. However, due to the big difference between the complexities of Local NN and Remote NN, the training is likely to be biased towards the Remote NN and ignore the Local NN's contribution, by assigning near-zero values to $\alpha$. Such bias could possibly make the training to be highly unstable or significantly reduce the inference accuracy, because the Local NN that perceives the top-$k$ important features may not be well trained.

\begin{figure}[ht]
	\centering
	\vspace{-0.1in}
	\hspace{-0.25in}
	\subfigure[Impact of T on $\alpha$] { 
		\includegraphics[width=0.24\textwidth]{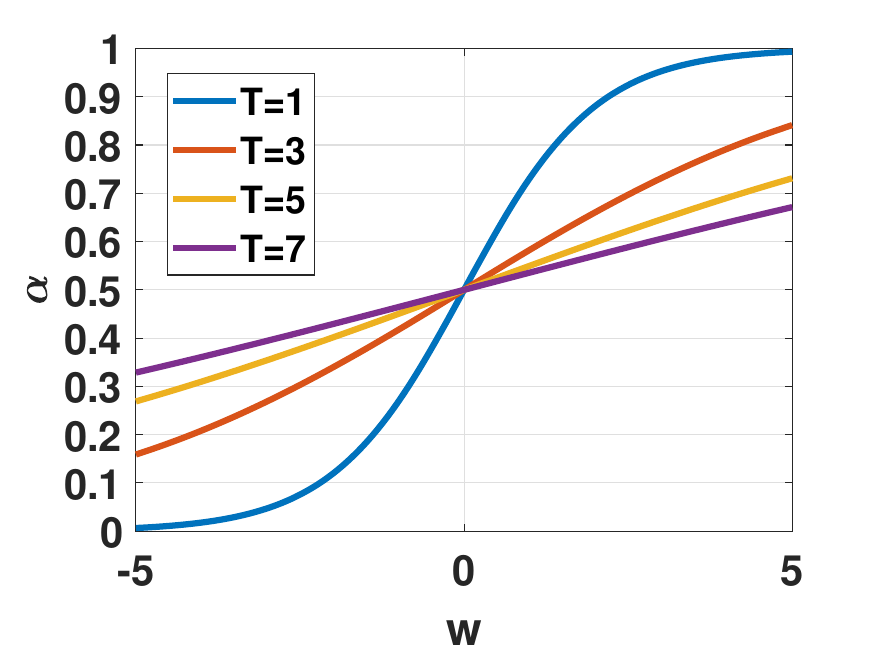}
		\label{fig:sigmoid_t}
	}
	\subfigure[Choices of T] { 
		\includegraphics[width=0.24\textwidth]{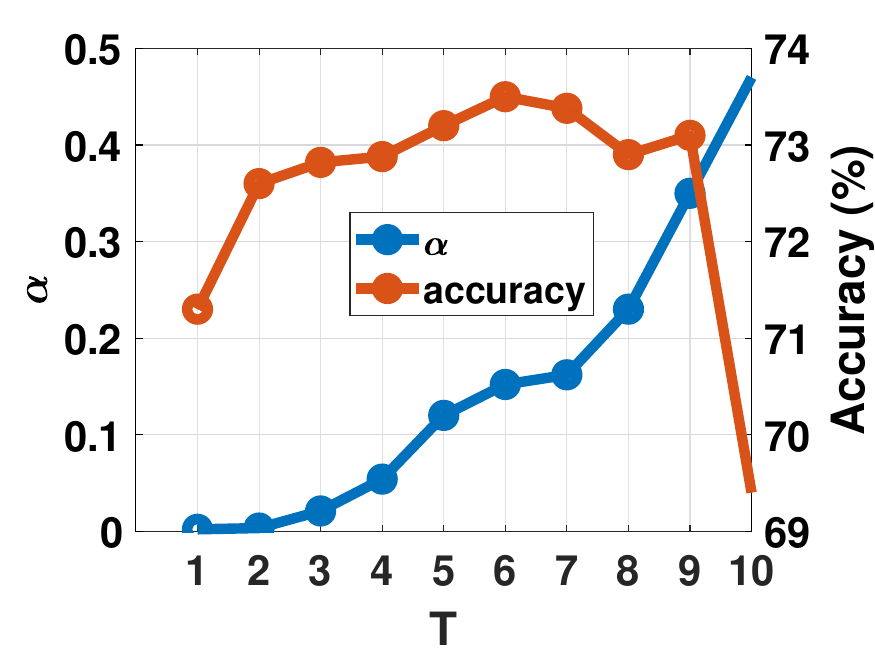}
		\label{fig:t_alpha_acc}
	}
	\hspace{-0.4in}
	\vspace{-0.1in}
	\caption{Prediction weighting with $\alpha$}
	\label{fig:prediction_weighting}
	\vspace{-0.1in}
\end{figure}

To avoid this bias, in AgileNN we introduce a soft constraint by formulating $\alpha$ as a parameterized sigmoid function:
\begin{equation*}
\alpha(w;T) = \frac{1}{1 + e^{-w/T}},
\end{equation*}
where $w$ is a trainable parameter and $T$ controls $\alpha$'s sensitivity to $w$. As shown in Figure \ref{fig:sigmoid_t}, the higher $T$ is, the slower $\alpha(w;T)$ varies along with $w$, and hence the less likely that the value of $\alpha$ will approach 0 or 1 during training. In practice, as shown in Figure \ref{fig:t_alpha_acc}, a moderate value of $T$ between 4 and 8 can effectively avoid biased values of $\alpha$ and ensure high inference accuracy.

The trained value of $\alpha$ is loaded to AgileNN runtime at the local device. In real-world settings, when the feature extractor does not correctly evaluate the importance of some features due to the possible inaccuracy in XAI, the user could flexibly fine-tune AgileNN's strategy of NN partitioning at run-time by reconfiguring the value of $\alpha$, to mitigate the loss of inference accuracy.

\section{Skewness Manipulation}
In this section, we provide technical details about how the training loss function in AgileNN's training is constructed based on the feature importances evaluated by XAI, so as to enforce the required skewness of such importances among the extracted features.


\subsection{Feature Ordering}
\label{subsec:feature_ordering}
Since XAI evaluation of feature importance builds on accumulating gradients in training and is hence unavailable during online inference, AgileNN makes sure that its feature extractor always generates the top-$k$ features with highest importance in the first $k$ channels in the output feature vector, as shown in Figure \ref{fig:feature_orders} - top, so that AgileNN runtime at the local embedded device can correctly identify them for every input data during inference.

\begin{figure}[ht!]
	\centering
	\hspace{-0.25in}
	\subfigure[Different feature orders] { 
		\includegraphics[width=0.23\textwidth]{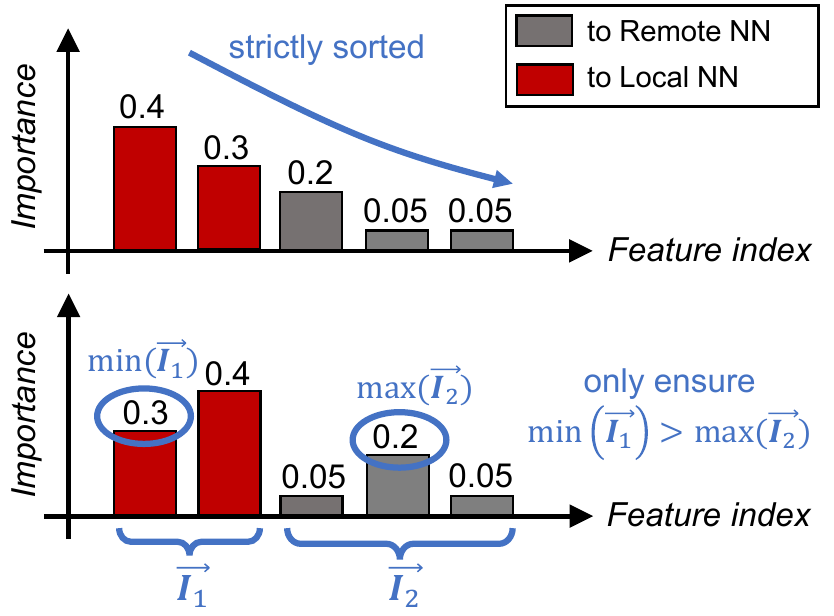}
		\label{fig:feature_orders}
	}
	\subfigure[Effectiveness of reordering] { 
		\includegraphics[width=0.23\textwidth]{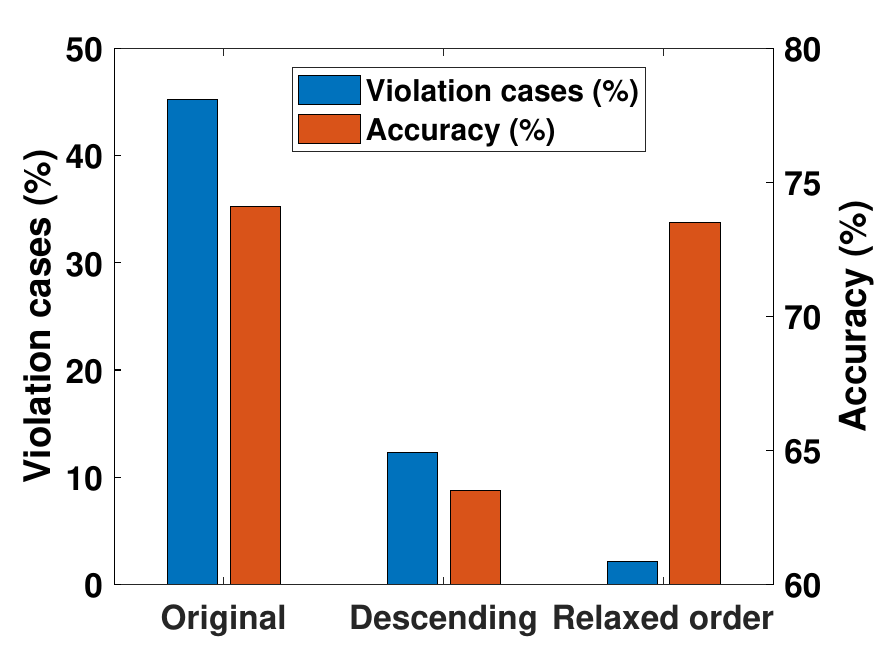}
		\label{fig:feature_shuffling_performance}
	}
	\hspace{-0.4in}
	\vspace{-0.1in}
	\caption{Feature reordering}
	\label{fig:feature_shuffling}
	\vspace{-0.1in}
\end{figure}

In training, a straightforward method to achieve this learning objective is to adopt the following loss function:
\begin{equation*}
L_{\text{descent}} = \|\vec{I}-\vec{I}_{\text{sorted}}\|_2^2,
\end{equation*}
where $\vec{I}$ denotes the normalized importances of the currently extracted features and $\vec{I}_{\text{sorted}}$ denotes the sorted form of $\vec{I}$ in the descending order. Minimizing this loss, hence, ensures that the extracted features are always sorted in the descending order of their importances. However, strictly enforcing such descending order in the produced feature vector requires high representation power in the feature extractor, or adds extra confusions in training if the feature extractor being used is too lightweight. To demonstrate this, we conduct preliminary experiments by using the feature extractor of the MobileNetV2 model \cite{sandler2018mobilenetv2} on the CIFAR-100 dataset \cite{krizhevsky2009learning}. As shown in Figure \ref{fig:feature_shuffling_performance}, enforcing such descending order in the output feature vector reduces the inference accuracy by $>$10\%.

Instead, we relax the learning objective by reducing the number of features being repositioned. As shown in Figure \ref{fig:feature_orders} - bottom, we do not require that all the features are sorted in the descending order of their importance, but instead only require that any top-$k$ feature's importance is higher than any other feature's importance. If any violation is found during training, a penalty will feedback to the NN for parameter update. Based on this relaxed learning objective, we construct our disorder loss as shown in Eq. (\ref{eq:disorder_loss}), which will only be non-zero if any violation of feature ordering occurs in training. In theory, this loss function of feature disordering is almost always differentiable \cite{nair2010rectified} and can be seamlessly incorporated into the regular training procedure\footnote{NNs are typically trained by providing gradient-based feedback being calculated from the loss function.}. As shown in Figure \ref{fig:feature_shuffling_performance}, $L_{\text{disorder}}$ can reduce the percentage of disorder cases to $<$2\% without impairing the inference accuracy.


\vspace{-0.05in}
\begin{figure}[h]
	\centering
	\includegraphics[width=0.8\linewidth]{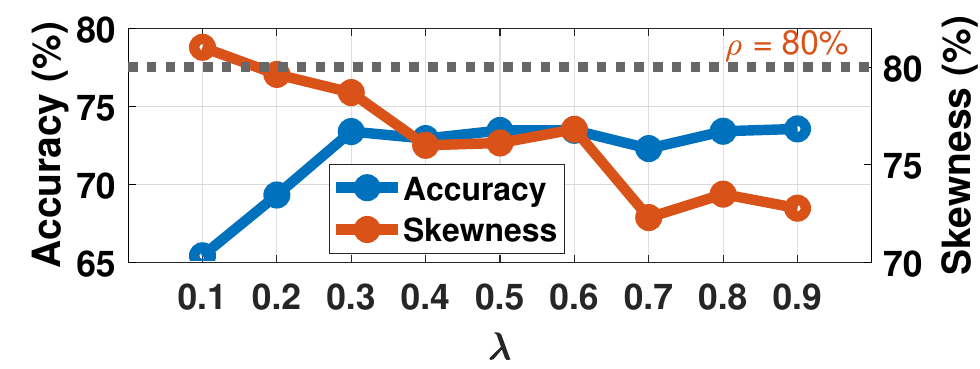}
	\vspace{-0.15in}
	\caption{Impact of $\lambda$ on CIFAR-100 dataset}
	\vspace{-0.1in}
	\label{fig:lambda_impact}
\end{figure}

\begin{algorithm}
	\caption{Selecting the $k$ initial feature channels}
	\label{alg:position_selection}
	\begin{algorithmic}[1]
		\REQUIRE $\mathbf{D}_{\text{train}}$: the training dataset with $N$ samples; \\
		$\mathcal{T}_{XAI}(\cdot)$: XAI-enabled Feature Importance Evaluator; \\
		$\mathcal{E}(\cdot)$: Feature extractor that outputs $C$ channels
		\ENSURE $(j_1, j_2, ..., j_k)$: The $k$ selected feature channels.
		\STATE{$(p_1, p_2, ..., p_C) \leftarrow 0$}
		\hfill\algorithmiccomment{\emph{initialize}} \\
		\FOR{\textbf{each } $d_i \in \mathbf{D}_{\text{train}}$} 
		\STATE {$F \leftarrow \mathcal{E}(d_i)$} \hfill\algorithmiccomment{ \emph{extract features}}
		\STATE {$I \leftarrow \mathcal{T}_{XAI}(F)$}
		\hfill\algorithmiccomment{\emph{evaluate feature importance}}
		\STATE{$F_{\text{sorted}} \leftarrow \text{sort}_{I}(F)$}
		\hfill\algorithmiccomment{\emph{sort features by their importance in descending order}}
		\STATE{$F_{top-k} \leftarrow F_{\text{sorted}}[1:k]$}
		\hfill\algorithmiccomment{\emph{extract the top-$k$ features with high importance}}
		\FOR{c = 1,...,C}
		\IF{$F[c] \in F_{top-k}$}
		\STATE{$p_c \leftarrow p_c + 1/N$}
		\ENDIF
		\ENDFOR
		\ENDFOR
		\STATE{$R \leftarrow \text{argsort}(p_1, p_2, ..., p_C)$} 
		\hfill\algorithmiccomment{\texttt{get the ranking of channels by their likelihood}}
		\STATE{$(j_1, j_2, ..., j_k) \leftarrow R[1:k]$}
		\hfill\algorithmiccomment{\texttt{decide top-$k$ channels}}
	\end{algorithmic}
	
\end{algorithm}

\subsection{Combined Training Loss}
\label{subsec:combined_loss}
To enforce the skewness requirement, we want that the cumulative normalized importance of top-$k$ features exceeds the given threshold $\rho$, and hence define the skewness loss as shown in Eq. (\ref{eq:skewness_loss}). Then, we combine the disorder loss and skewness loss together to construct the training loss for skewness manipulation as:
\begin{equation*}
L = \lambda \cdot L_{\text{prediction}} + (1-\lambda) \cdot (L_{\text{skewnss}} + L_{\text{disorder}})
\end{equation*}
where $L_{\text{prediction}}$ is the standard prediction loss and $\lambda$ is a hyperparameter within $(0, 1)$ to control the contributions of $L_{\text{skewnss}}$ and $L_{\text{disorder}}$ in training feedback. In practice, according to our preliminary results in Figure \ref{fig:lambda_impact}, aggressively reducing $\lambda$, although achieving higher skewness, could reduce the impact of prediction loss in training feedback and hence impair the NN inference accuracy. In contrast, we observe that a moderate value of $\lambda$ between 0.2 and 0.4 could effectively approximate to the skewness requirement with the minimum impact on NN inference accuracy. Alternatively, one can also adopt the techniques on NN loss balancing \cite{chen2018gradnorm, groenendijk2021multi} for adaptive adjustment of $\lambda$ at runtime.


\begin{figure}[ht]
	\centering
		\vspace{-0.1in}
	\includegraphics[width=0.27\textwidth]{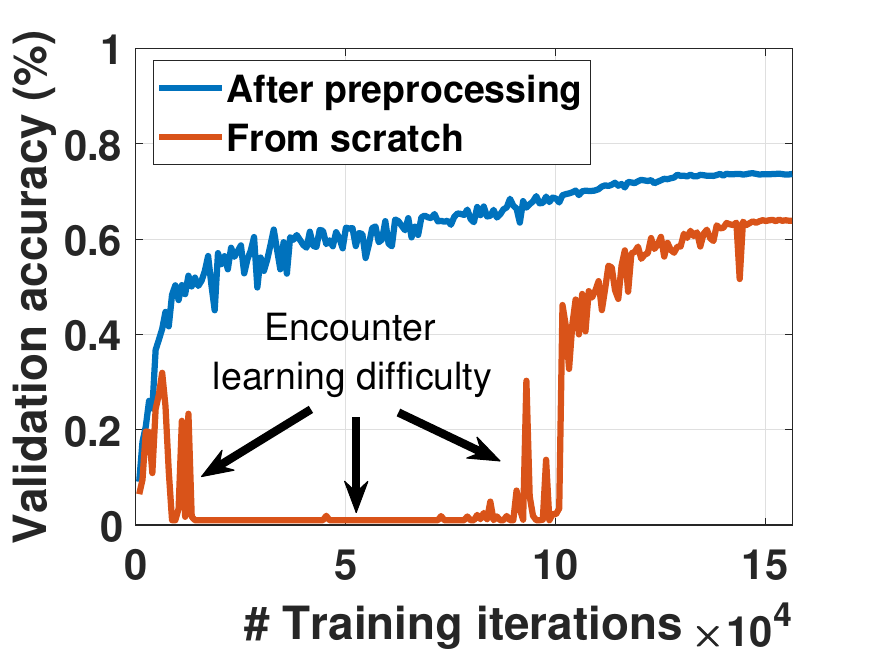}
	\vspace{-0.1in}
	\caption{Effectiveness of Pre-processing}
	\label{fig:preprocessing_acc_curve}
	\vspace{-0.15in}
\end{figure}

\section{Pre-processing the Feature Extractor}
In this section, we describe in detail how we pre-process the feature extractor by selecting the initial $k$ feature channels as the input to the Local NN in joint training. Intuitively, we can randomly select $k$ feature channels and mandate the feature extractor to produce the top important features in these channels using the disorder loss described in Section 3.1. However, such arbitrary selection will bring serious learning difficulty that leads to low training quality. We demonstrate this by doing preliminary experiments on the CIFAR-100 dataset with such random channel selection. As shown in Figure \ref{fig:preprocessing_acc_curve}, the NN experiences learning difficulty from the beginning epochs and it eventually causes poor convergence.

Instead, we make such channel selection based on the likelihood that one of top-$k$ features with high importance is located in a channel, and compute such likelihood from the training data. More specifically, as described in Algorithm \ref{alg:position_selection}, such likelihood of each channel is cumulatively computed from all the $N$ data samples in the training dataset, and increases by $1/N$ every time when a data sample's top-$k$ features with high importance are located in the channel. As shown in Figure \ref{fig:preprocessing_acc_curve}, our pre-processing can largely reduce the learning difficulty and ensure the quality of training.

\begin{figure}[ht]
	\centering
	\vspace{-0.1in}
	\includegraphics[width=0.36\textwidth]{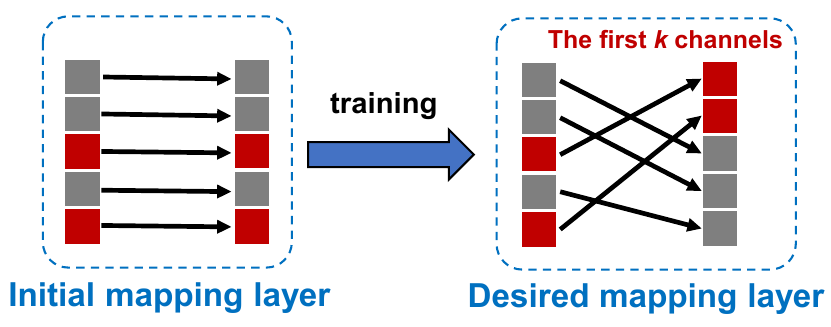}
	\vspace{-0.1in}
	\caption{Training the mapping layer}
	\label{fig:mapping_layer}
	\vspace{-0.1in}
\end{figure}

After having selected these initial $k$ channels, we expect the joint training process will be able to gradually enforce the required feature ordering, as described in Section \ref{subsec:feature_ordering}, through the disorder loss. To facilitate this, as shown in Figure \ref{fig:mapping_layer}, in AgileNN's training we add an extra mapping layer between the feature extractor and the Local NN, and instruct the training process to ensure that the top-$k$ important features reside in the first $k$ channels of the output feature vector. After the training finishes, this mapping layer will be discarded and only the feature extractor is used in inference.


\begin{figure}[ht]
	\centering
	\includegraphics[width=0.6\linewidth]{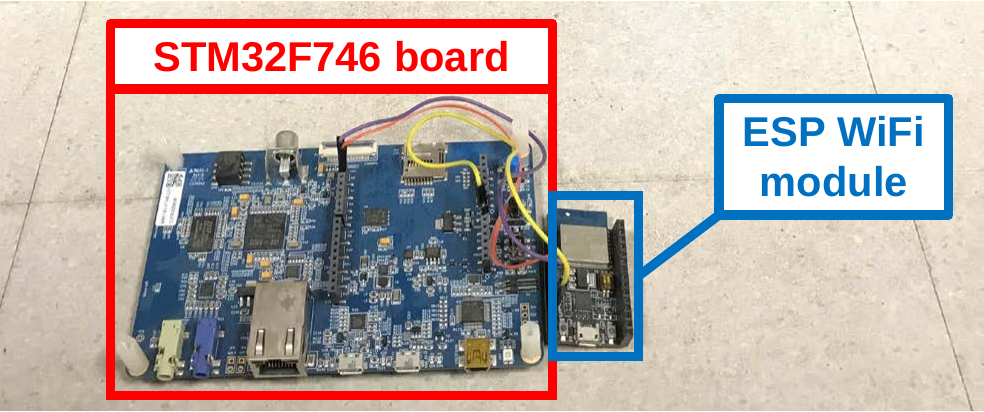}
	\vspace{-0.1in}
	\caption{Devices in our implementation}
	\vspace{-0.1in}
	\label{fig:device_layout}
\end{figure}

\section{Implementation}
As shown in Figure \ref{fig:device_layout}, we use a STM32F746NG MCU board\footnote{https://www.st.com/en/microcontrollers-microprocessors/stm32f746ng.html} as the local embedded device, which is widely used as the computing platform in current tinyML and on-device AI research (e.g., MCUNet \cite{lin2020mcunet}). It is equipped with an ARM 32-bit Cortex-M7 CPU at 216MHz, 320KB SRAM and 1MB flash storage, and supports flexible CPU frequency scaling to provide different amounts of on-device computing power. In addition, since neural network inference on the Cortex M series of MCUs has been officially supported by the TensorFlow community\footnote{https://www.tensorflow.org/lite/microcontrollers}, we believe that using these MCUs to implement and evaluate AgileNN could better justify AgileNN's practical merits, compared to using other MCUs such as the MSP430 series.

The MCU board uses an ESP-WROOM-02D WiFi module to transmit data to a server. The server is a Dell Precision 7820 workstation that equips with a 3.6GHz 8-core Intel Xeon CPU, 128GB main memory and an Nvidia RTX A6000 GPU with 48GB memory.

As shown in Figure \ref{fig:implementation}, our offline training in AgileNN is implemented using TensorFlow Python library, and we converted the Local NN from a \texttt{float32} model into an \texttt{int8} model using TensorFlow Lite Converter. This model is then casted to a static binary array for better memory efficiency on the local device. We use TF Micro runtime to execute the \texttt{int8} model on the STM32 board. To further reduce the Local NN's computing latency, we merge the original TF Micro runtime with CMSIS-NN 5.0, which provides extra acceleration on several selected NN operations on ARM devices. On the other hand, the Remote NN remains full precision and is executed by TensorFlow Python runtime on the server.

\begin{figure}[ht]
	\centering
	\vspace{-0.05in}
	\includegraphics[width=\linewidth]{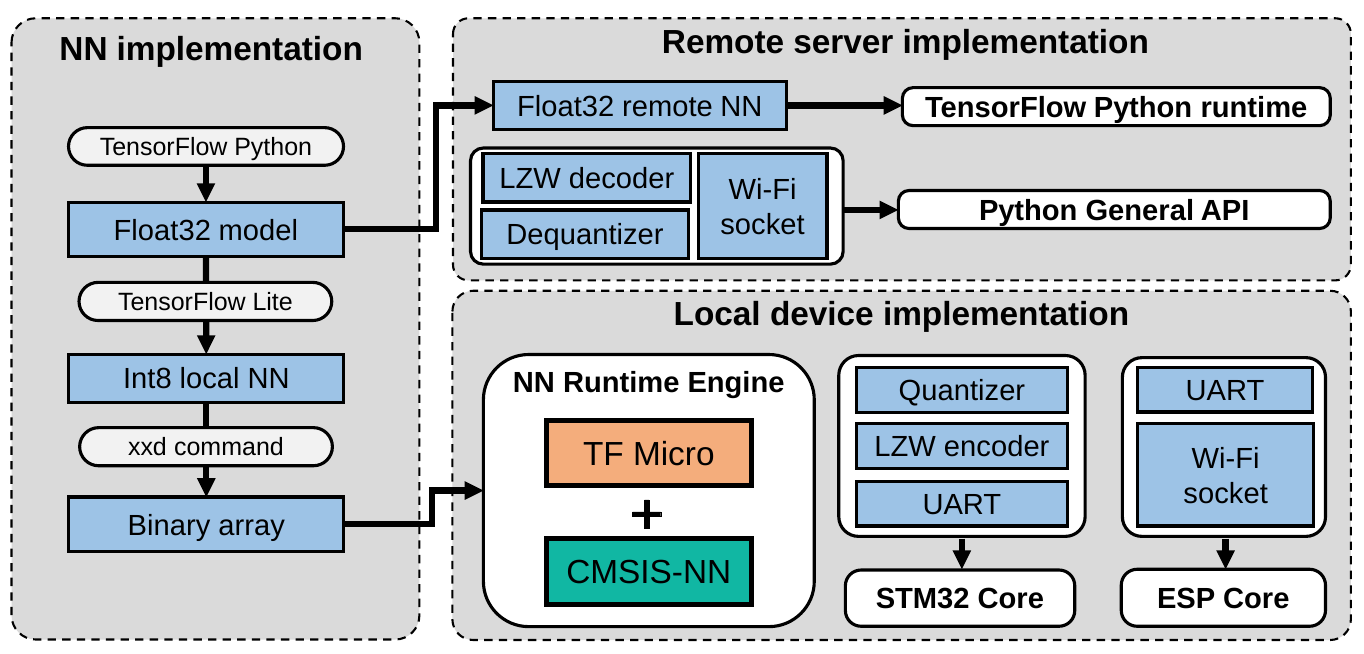}
	\vspace{-0.2in}
	\caption{AgileNN implementation}
	\vspace{-0.05in}
	\label{fig:implementation}
\end{figure}

On the STM32F746 board, we use STM32CubeIDE to implement its software in C++ and configure the embedded hardware. To compress the less important features before transmission, we first adopt learning-based quantization \cite{agustsson2017soft} and then apply standard LZW compression \cite{nelson1989lzw}. The compressed features are delivered to the WiFi module through UART, and the WiFi module transmits these features to the server through a UDP link at 6 Mbps.

On the server, we write a custom Python script to communicate with the STM32F746 board via general socket APIs, and verifies the integrity of the received features being applying them to the Remote NN. 



\section{Performance Evaluation}
In our evaluations, to meet the embedded device's local resource constraints, we construct AgileNN's feature extractor with two convolutional layers, each of which has 24 output channels. The Local NN in AgileNN has the minimum complexity, and contains one global-average pooling layer and one dense layer. The Remote NN in AgileNN is constructed by removing the first convolutional layer from the MobileNetV2 \cite{sandler2018mobilenetv2} model. In all the evaluations, the sizes of feature extractor, Local NN and Remote NN in AgileNN remain fixed, but we vary the compression rate when transmitting the set of less important features from local to remote.

We evaluate AgileNN over multiple datasets listed below, and scaled all images in datasets to 96x96 in our experiments. Due to the low memory capacity of the embedded device, we focus on image recognition tasks instead of memory-demanding learning tasks, such as audio and video analytics \cite{amodei2016deep, wu2015modeling} that require expensive preprocessing steps \cite{li2011reducing}. 
\begin{itemize}
	\item \textbf{CIFAR-10/100 \cite{krizhevsky2009learning}}: This dataset contains 50k training images and 10k testing images that belong to 100 different categories and 10 super categories. 
	\item \textbf{SVHN \cite{netzer2011reading}}: This dataset contains 73k training images and 26k testing images about street address numbers. 
	\item \textbf{ImageNet-200 \cite{le2015tiny}}: This is a subset of ImageNet dataset \cite{deng2009imagenet} that contains 100k training images and 10k testing images that are classified into 200 categories.
\end{itemize}

In training, AgileNN adopts an EfficientNetV2 CNN \cite{tan2021efficientnetv2} that is pre-trained on the ImageNet dataset as the reference network, and the training hyperparameters are configured the same as MobileNetV2's setting \cite{sandler2018mobilenetv2}. We use the SGD optimizer with a learning rate of 0.1, and the standard weight decay is set to $5\times10^{-4}$ and all the training runs for 200 epochs. The batch size in training is set to be 128 for the CIFAR-10 dataset and 64 for all other datasets.

In our evaluations, all the experiment results are averaged over the entire testing dataset. We compare AgileNN with the baseline of edge-only inference and three existing approach NN inference approaches, which span both categories of local inference and NN partitioning:
\begin{itemize}
	\item \textbf{Edge-only inference}: The entire local data is compressed by the LZW compressor and transmitted to the server for inference.
	\item \textbf{MCUNet \cite{lin2020mcunet}}: The entire NN is running at the local embedded device, and the NN structure is optimally discovered by NAS according to the on-device resource constraint on the NN complexity.
	\item \textbf{DeepCOD \cite{yao2020deep}}: A NN-based encoder is embedded on the local device to transform the raw data or features into a more compressible form. The encoder is trained with the sparsity constraint in an end-to-end manner\footnote{AgileNN is equivalent to DeepCOD \cite{yao2020deep} if the top-$k$ features with high importance are also compressed and sent to the server.}.
	\item \textbf{SPINN \cite{laskaridis2020spinn}}: Besides NN partitioning, early-exit structures are incorporated in the NN to adaptively adjust the NN complexity for runtime inference.
\end{itemize}

In particular, since MCUNet's NN design adopts different input resolutions for different datasets, we make sure to always use the same image resolution among all other approaches, including AgileNN, to make fair comparisons among different approaches.

\begin{figure}[ht!]
	\centering
	\vspace{-0.05in}
	\hspace{-0.25in}
	\subfigure[Test accuracy] { 
		\includegraphics[width=0.22\textwidth]{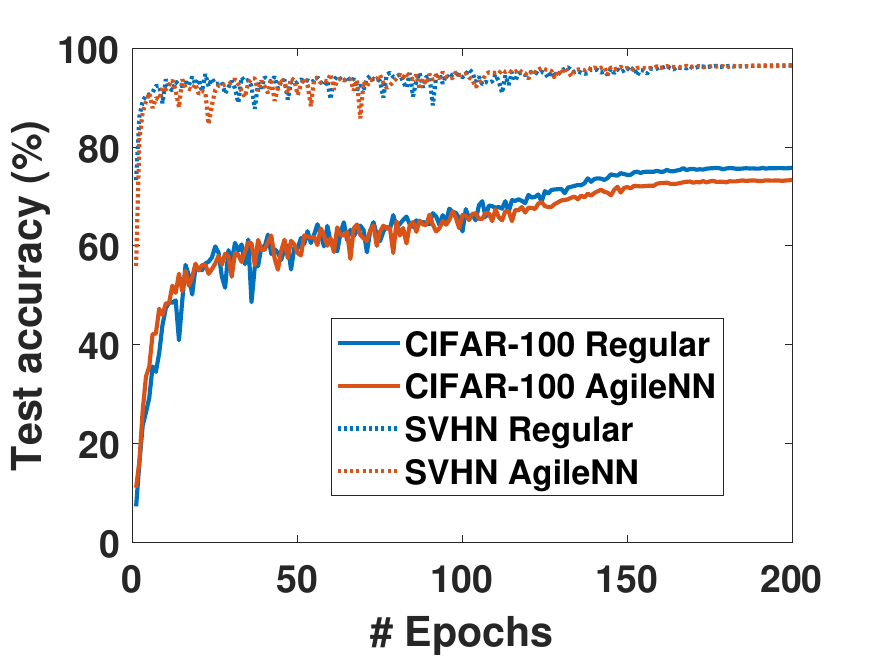}
		\label{fig:test_accuracy_epoch_curve_cifar100_svhn}
	}
	\subfigure[Test loss] { 
		\includegraphics[width=0.22\textwidth]{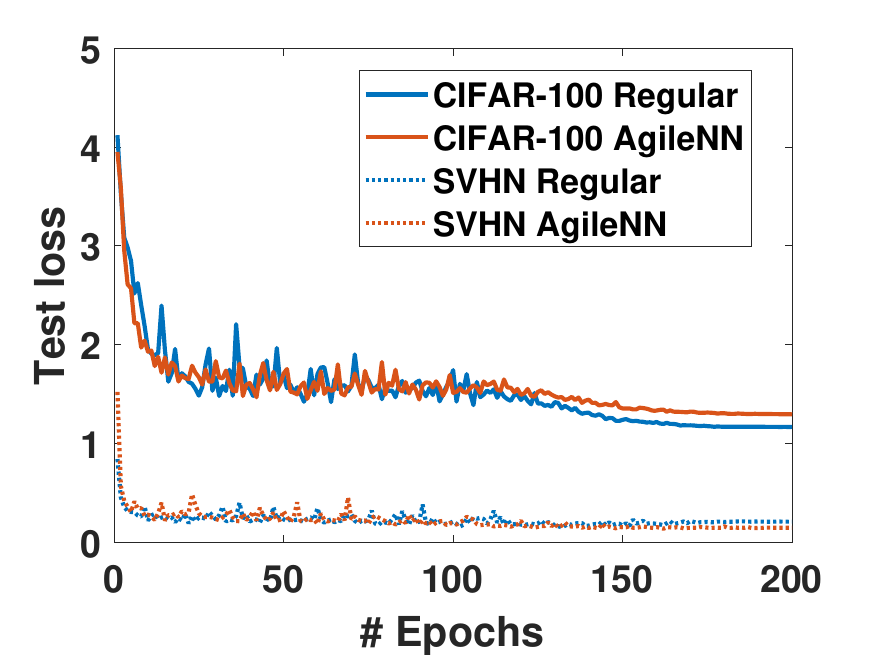}
		\label{fig:test_loss_epoch_curve_cifar100_svhn}
	}
	\hspace{-0.4in}
	\vspace{-0.1in}
	\caption{AgileNN's training performance on CIFAR-100 and SVHN datasets}
	\label{fig:training_convergence}
	\vspace{-0.1in}
\end{figure}

\subsection{Training Convergence and Cost}
As a prerequisite, we first evaluate the quality and cost of AgileNN's training. As shown in Figure \ref{fig:training_convergence}, during the training procedure, AgileNN exhibits a very similar rate of training convergence, in terms of test accuracy and loss, compared to regular training of MobileNetV2 on CIFAR-100 and SVHN datasets. These results show that, although the added feature ordering and skewness manipulation increases the learning complexity, AgileNN can still ensure fast training convergence with the appropriate loss function design and preprocessing of the feature extractor.

On the other hand, with the extra computations of feature importance using XAI and the corresponding involvement of extra training feedback, we observe 3x-4x wall-clock time increase for each training epoch in AgileNN. However, since the training of feature extractor, Local and Remote NNs is conducted offline, such time increase will not affect AgileNN's online performance on weak embedded devices. Reduction of such training time can be done by either using stronger computing hardware (e.g., stronger GPUs) or more lightweight XAI tools \cite{ShrikumarGK17,hesse2021fast}.


\subsection{Accuracy and Latency of NN Inference}

In general, the accuracy of NN inference can be improved by using more complicated NNs, which in turn result in longer inference latency. For easier comparisons, we configure the existing approaches' NN complexities so that the difference between their's and AgileNN's inference accuracy is always within 10\%. In these cases, we compare the AgileNN's end-to-end inference latency with theirs. Note that for local inference approaches such as MCUNet, the inference latency is only determined by the local NN computing time. For NN partitioning approaches including DeepCOD, SPINN and AgileNN, the inference latency consists of 1) the local NN computing time, 2) the local computing time for data compression, 3) the network transmission time and 4) the remote NN time for data decompression and computing.

\begin{figure}[ht]
	\centering
	\vspace{-0.1in}
	\subfigure[CIFAR-10] { 
		\includegraphics[width=0.22\textwidth]{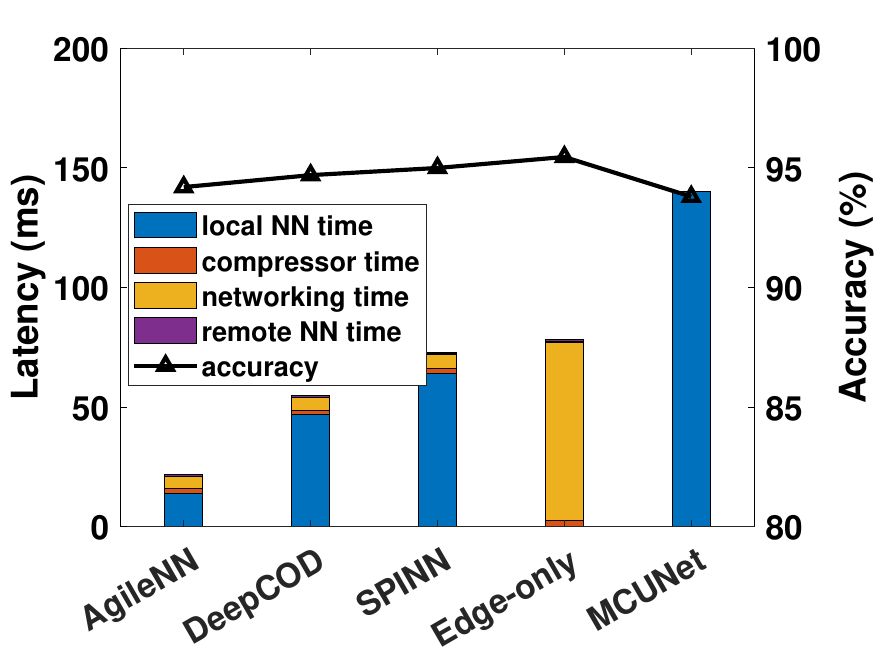}
		\label{fig:cifar10_latency_acc}
	}
	\subfigure[CIFAR-100] { 
		\includegraphics[width=0.22\textwidth]{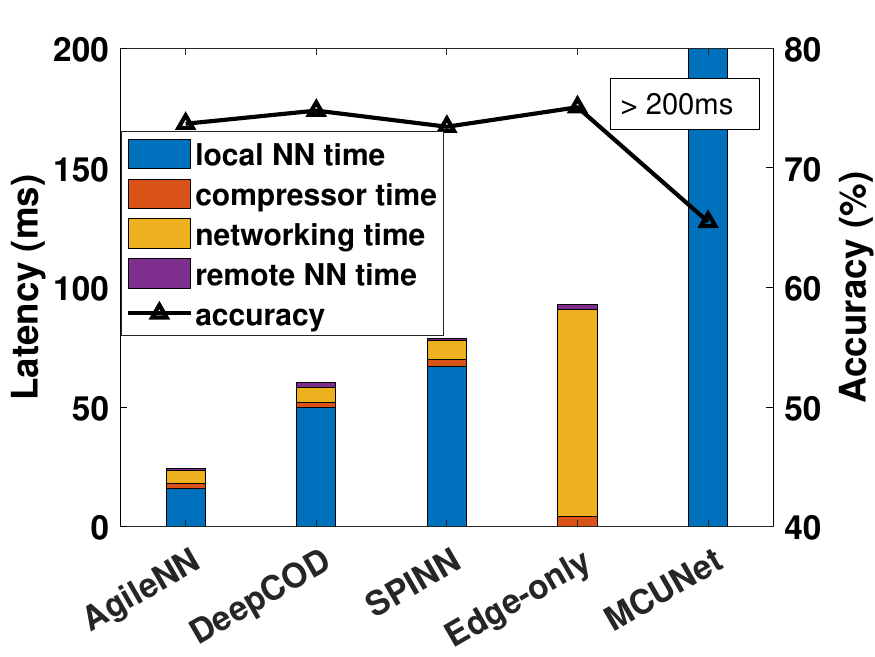}
		\label{fig:cifar100_latency_acc}
	}\newline
	
	\subfigure[SVHN] { 
		\includegraphics[width=0.22\textwidth]{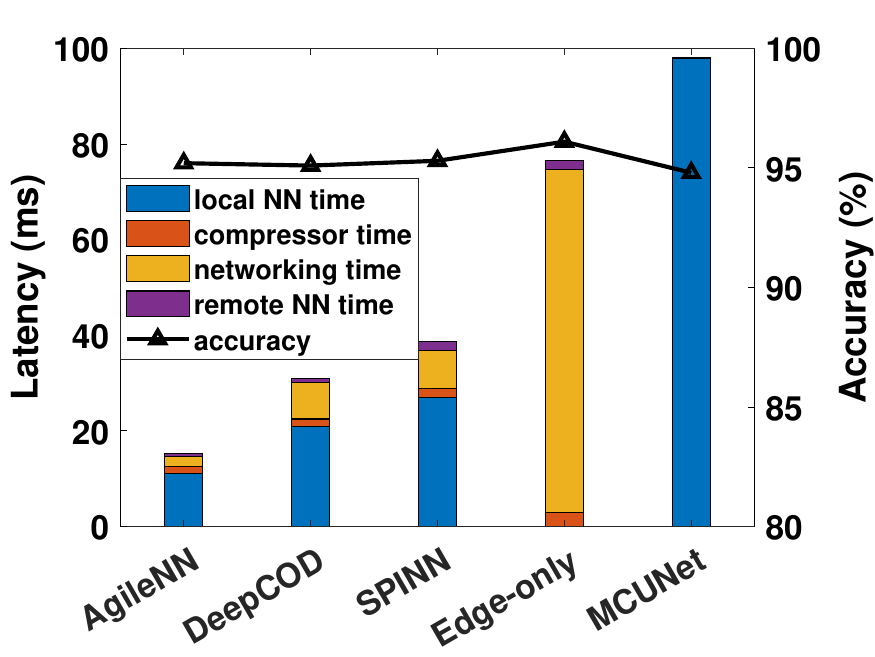}
		\label{fig:svhn_latency_acc}
	}
	\subfigure[ImageNet-200] { 
		\includegraphics[width=0.22\textwidth]{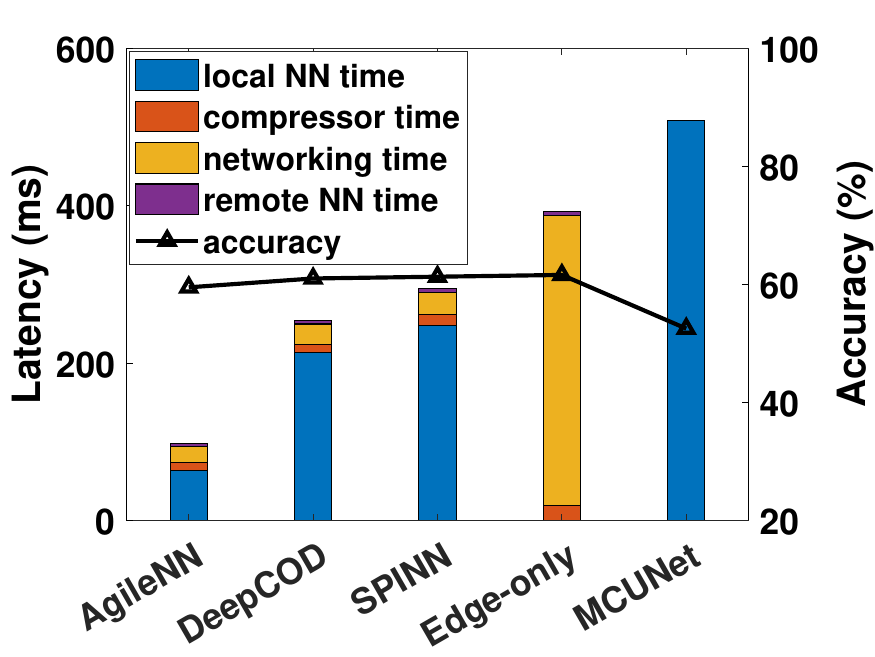}
		\label{fig:imagenet200_latency_acc}
	}
	
	\hspace{-0.4in}
	\vspace{-0.15in}
	\caption{Latency and accuracy of NN inference}
	\label{fig:latency_acc}
\end{figure}

Results in Figure \ref{fig:latency_acc} show that, AgileNN is able to reduce the end-to-end inference latency by 2x-2.5x when compared to all the existing approaches, while retaining similar inference accuracy with DeepCOD and SPINN. In particular, such latency in most datasets can be effectively controlled within 20ms, which is comparable to the sampling interval of many embedded sensor devices\footnote{For example, most camera sensors on embedded devices have a sampling rate of 30Hz during video capture. The sampling rate of environmental sensors (e.g., temperature sensors) is usually $<$10Hz due to the slower changes of the physical environment \cite{bakker1999low}. The sampling rate of IMU sensors is usually capped at 100Hz, but a lower rate is used more often in practice to save power \cite{diaz2013optimal}.}. As a result, AgileNN can effectively support real-time NN inference on weak embedded devices, by ensuring sure that generated sensory data can always be timely consumed. 

More specifically, Figure \ref{fig:latency_acc} shows that the AgileNN's latency reduction mainly comes from the lower local NN computing time, which can be reduced by up to 10x. Compared to DeepCOD and SPINN which have to use a complicated Local NN to ensure feature sparsity, the adoption of XAI in AgileNN allows achieving higher feature sparsity with a much more lightweight Local NN and feature extractor. The inference latency of MCUNet is much higher (100-500ms) than that of other approaches, due to the complicated NN that is fully executed on the embedded device.

On the other hand, although edge-only inference incurs the minimum local computing delay, it suffers from the low wireless link rate at the local device\footnote{Due to the local resource constraint, the maximum WiFi data rate at the STM32F746 MCU's WiFi module is capped at 6 Mbps.} that results in a significantly higher wireless transmission latency due to the low data compressibility. The overall end-to-end latency of edge-only inference, hence, is higher than DeepCOD, SPINN and AgileNN.

\begin{table}[h]
	\begin{tabular}{c||c|c|c|c}
		\hline
		Dataset                     & CIFAR-10 & CIFAR-100 & SVHN   & ImageNet \\ \hline		\hline
		Reduction & 43.7\%   & 15.8\%    & 72.3\% & 20.8\%       \\ \hline
	\end{tabular}
	\vspace{0.05in}
	\caption{Reduction of transmitted data size, compared to DeepCOD \cite{yao2020deep}}
	\label{table:transmission_time_reduction}
	\vspace{-0.2in}
\end{table}

Such higher feature sparsity, on the other hand, also results in significant reduction on network transmission time. As shown in Table \ref{table:transmission_time_reduction}, such reduction on some datasets such as SVHN could exceed 70\%. This reduction, even being lower than 20\%, could be important in some IoT scenarios, where IoT devices are wirelessly connected to the 5G backbone network and will hence need to make usage-based payments to the 5G service provider \cite{wang2018iot}.

\begin{figure}[ht!]
	\centering
	\vspace{-0.15in}
	\hspace{-0.25in}
	\subfigure[CIFAR-100] { 
		\includegraphics[width=0.22\textwidth]{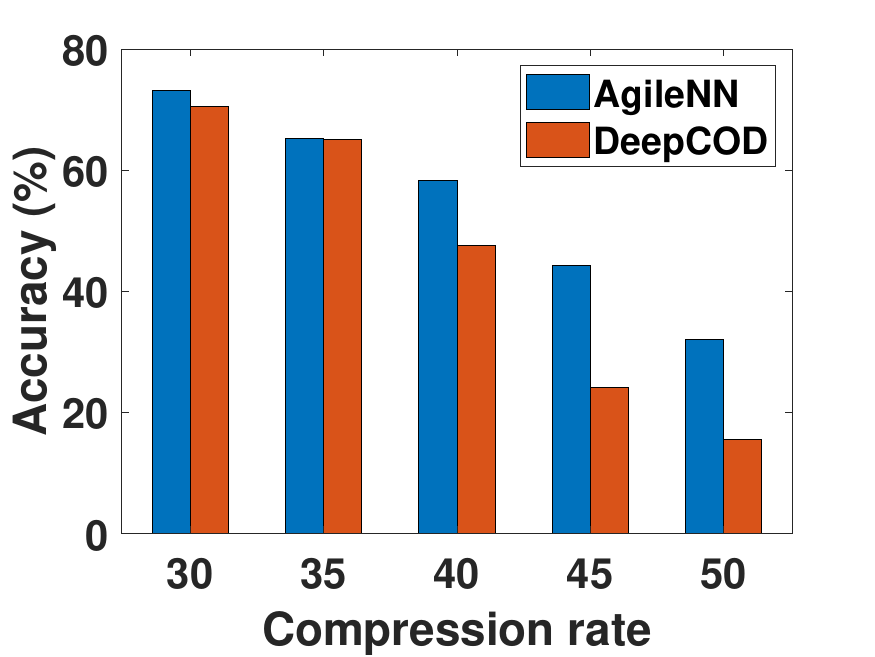}
		\label{fig:cifar100_cr_acc}
	}
	\subfigure[SVHN] { 
		\includegraphics[width=0.22\textwidth]{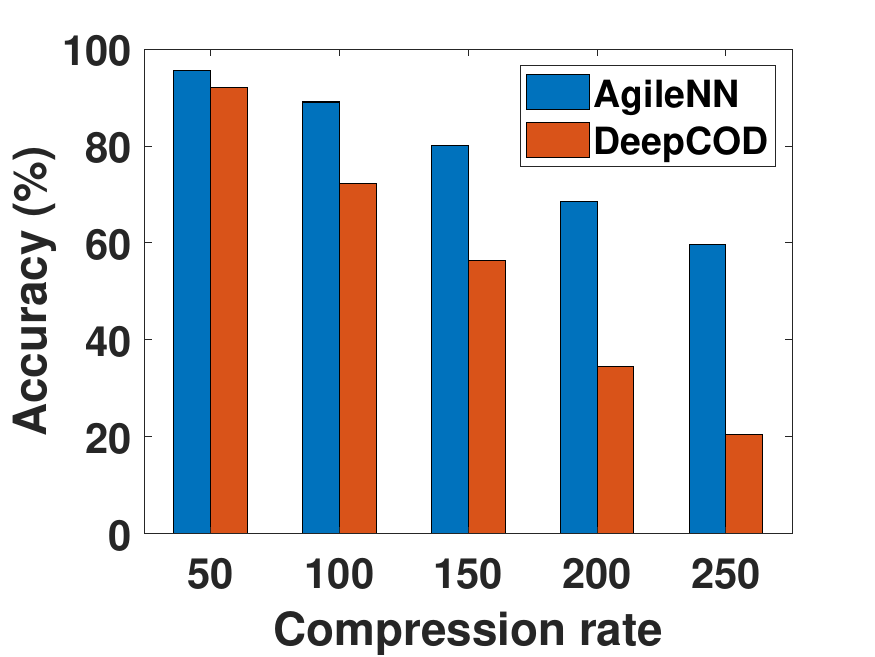}
		\label{fig:svhn_cr_acc}
	}
	\hspace{-0.4in}
	\vspace{-0.15in}
	\caption{Accuracy with different compression rates}
	\label{fig:cr_acc}
	\vspace{-0.05in}
\end{figure}

\noindent\textbf{The Impact of Compression Rate.} Since DeepCOD performs best among the three existing approaches for comparison, we further compare its performance with AgileNN when we apply different compression rates to transmit data features to the remote server. Results in Figure \ref{fig:cr_acc} over the CIFAR-100 and SVHN datasets show that, AgileNN can always achieve higher NN inference accuracy with the same compression rate being applied, due to its more agile and efficient enforcement of feature sparsity that results in better compressibility. In particular, when very high compression rates are applied, DeepCOD experiences significant accuracy reduction due to the limited representation power of its encoder, but such reduction in AgileNN is much lower.


\noindent\textbf{The Impact of Prediction Reweighting.} As described in Section 3.3, the predictions made by Local NN and Remote NN are combined towards the inference output, using a tuneable parameter $\alpha$. Results in Figure \ref{fig:reweighting} on the CIFAR-100 and SVHN datasets show that, the NN inference accuracy will significantly drop if highly biased values of $\alpha$ (e.g., close to 0 or 1) are being used. This is because using a very small $\alpha$ reduces the contribution of important features and could hence miss key information to inference. Increasing the value of $\alpha$, on the other hand, imposes majority of the inference task to the Local NN, which may  not be complicated enough to achieve high inference accuracy.

\begin{figure}[ht!]
	\centering
	\vspace{-0.2in}
	\hspace{-0.25in}
	\subfigure[CIFAR-100] { 
		\includegraphics[width=0.22\textwidth]{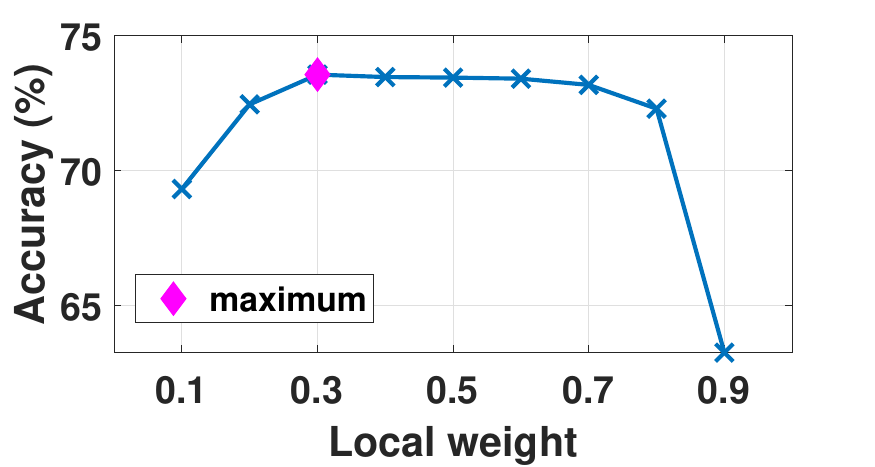}
		\label{fig:cifar100_weight}
	}
	\subfigure[SVHN] { 
		\includegraphics[width=0.22\textwidth]{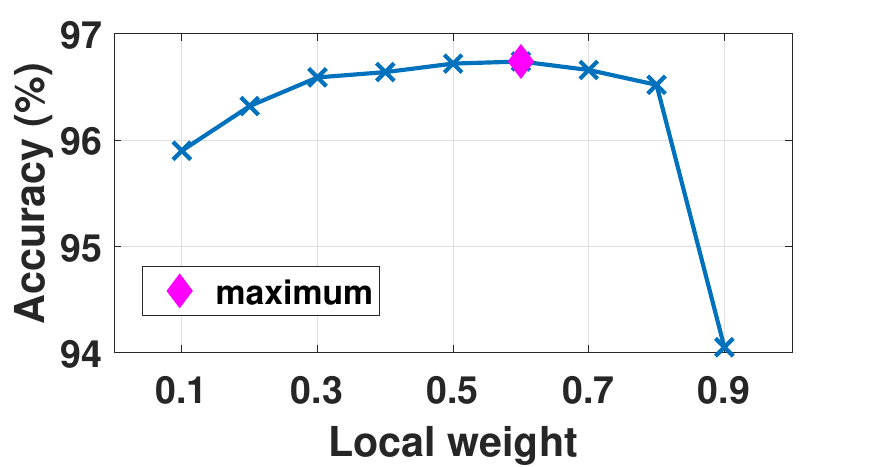}
		\label{fig:svhn_weight}
	}
	\hspace{-0.4in}
	\vspace{-0.2in}
	\caption{Applying different weights}
	\label{fig:reweighting}
	\vspace{-0.1in}
\end{figure}

Instead, we conclude that the maximum inference accuracy can be achieved when $\alpha=$0.3 for CIFAR-100 dataset and $\alpha=$0.6 for SVHN dataset. Note that the optimal value of $\alpha$ is dependent on the data characteristics in the training dataset. In practice, this value can either be jointly trained offline with the feature extractor and NNs, or be manually tuned online based on the specific data characteristics for better inference accuracy.


\subsection{Local Resource Consumption}
In this section, we evaluate the amount of local resources at the embedded device that are consumed by AgileNN's inference. Such local resources include 1) the local battery power and 2) the local memory and flash storage. For fair comparison between different schemes, being similar with the previous experiments, we keep the gap between different schemes' inference accuracy to be within 5\%.

\begin{figure}[ht]
	\centering
	\vspace{-0.2in}
	\hspace{-0.25in}
	\subfigure[CIFAR-100] { 
		\includegraphics[width=0.22\textwidth]{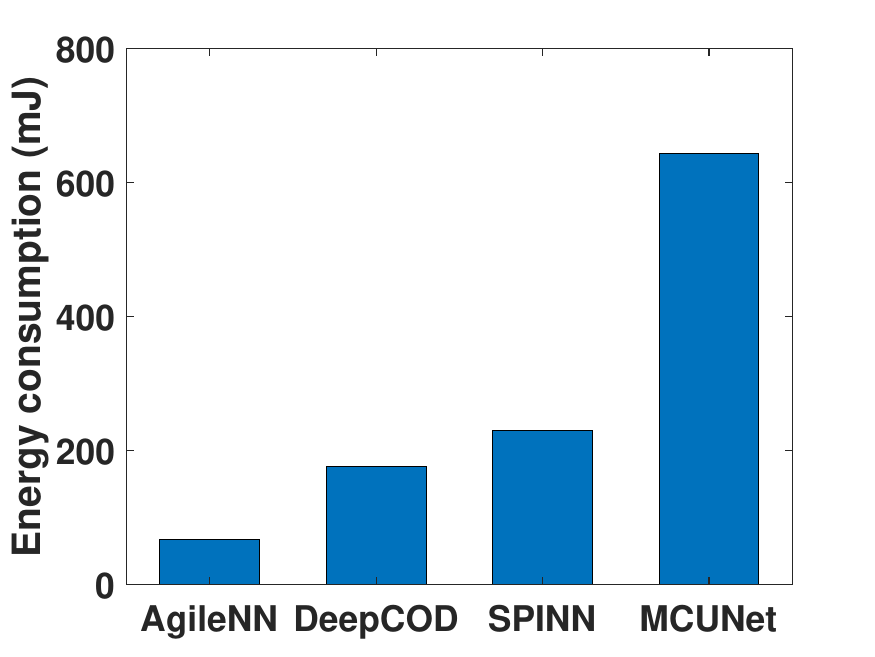}
		\label{fig:cifar100_energy_consumption}
	}
	\subfigure[ImageNet-200] { 
		\includegraphics[width=0.22\textwidth]{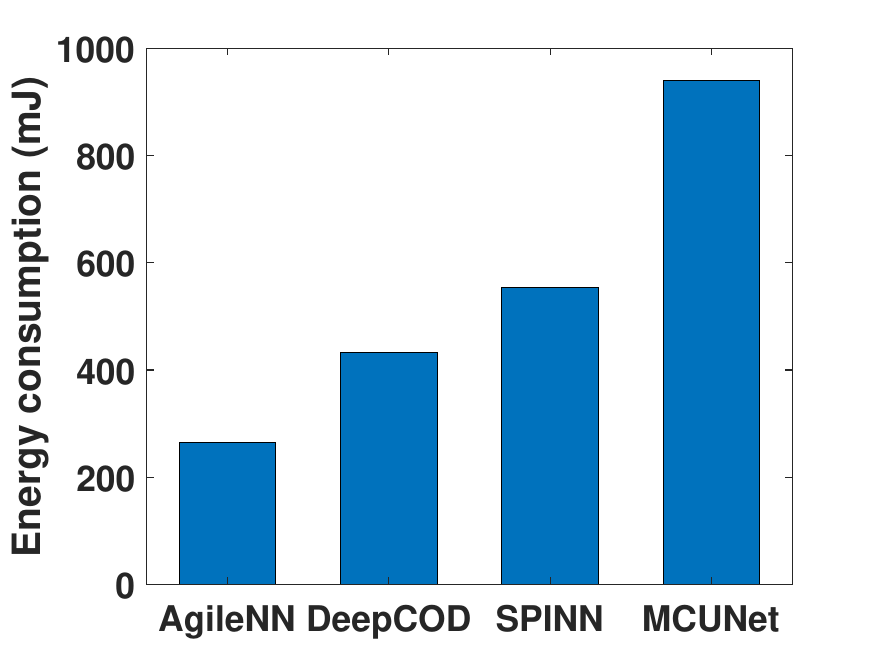}
		\label{fig:imagenet200_energy_consumption}
	}
	\hspace{-0.4in}
	\vspace{-0.1in}
	\caption{Local energy consumption per NN inference run}
	\label{fig:energy_consumption}
	\vspace{-0.1in}
\end{figure}

\noindent\textbf{Energy consumption.} We measure the amount of local device's energy consumption per NN inference run as the average over 100 inference runs. Such energy consumption includes both the local NN computing cost and data transmission cost via WiFi. As shown in Figure \ref{fig:energy_consumption}, since AgileNN uses a very lightweight feature extractor and local NN but achieves even higher feature sparsity with these lightweight NN structures, its runtime consumes less local energy in both computation and communication, leading to significantly higher energy efficiency. Especially when being used on smaller datasets such as CIFAR-100, its energy efficiency is at least 2.5x higher than that of DeepCOD, and is $>$8x higher than that of MCUNet.

\begin{figure}[ht]
	\centering
	\hspace{-0.25in}
	\subfigure[CIFAR-100] { 
		\includegraphics[width=0.22\textwidth]{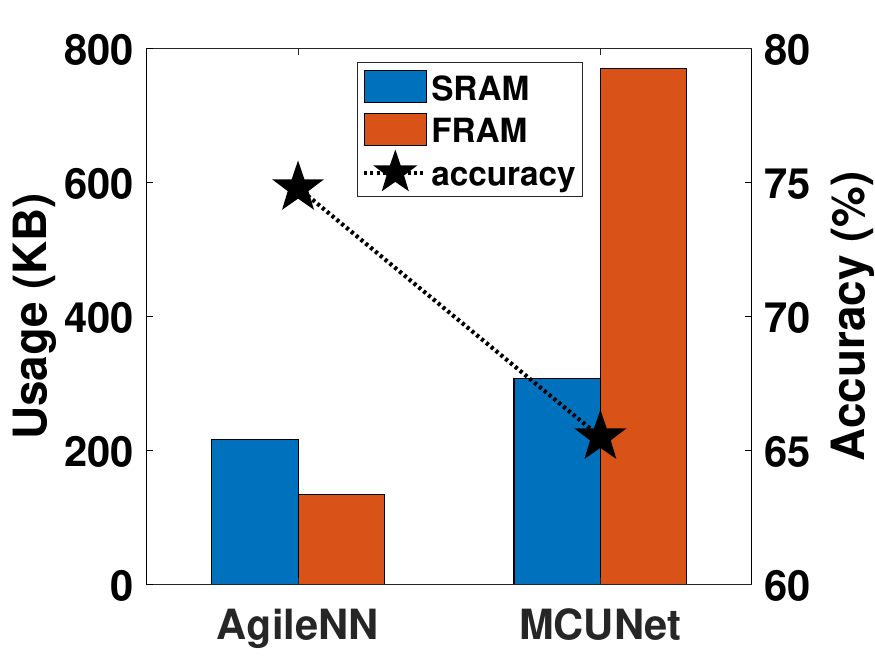}
		\label{fig:cifar100_sram_fram_acc}
	}
	\subfigure[ImageNet-200] { 
		\includegraphics[width=0.22\textwidth]{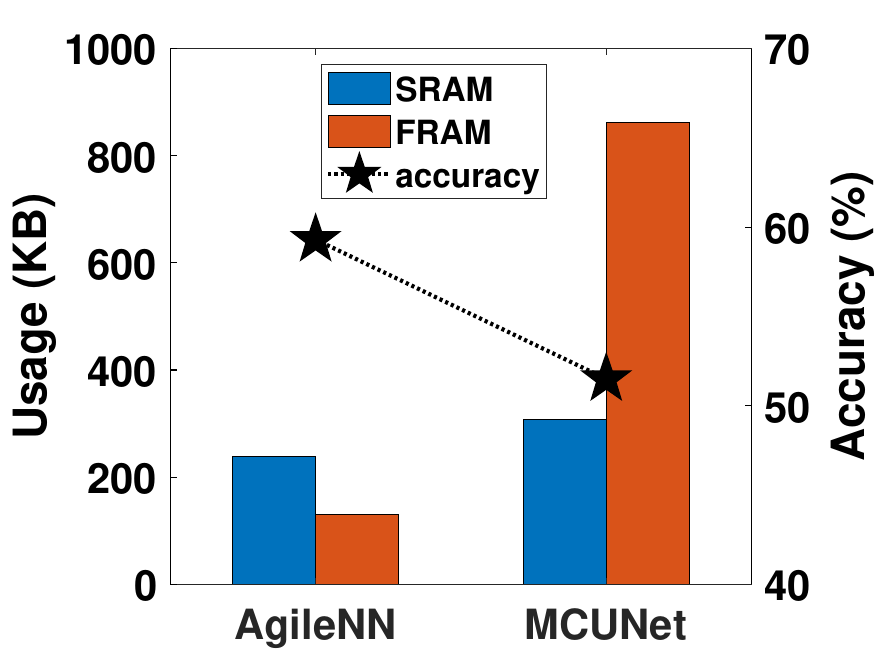}
		\label{fig:imagenet200_sram_fram_acc}
	}
	\hspace{-0.4in}
	\vspace{-0.15in}
	\caption{Memory and storage usage}
	\label{fig:sram_fram}
	\vspace{-0.15in}
\end{figure}

\noindent\textbf{Memory and storage usage.} We measure the usage of on-board memory (SRAM) and storage (FRAM) by using the STM32Cube debugging software. As shown in Figure \ref{fig:sram_fram}, due to the low complexity of feature extractor and NN, AgileNN's consumptions of the local device's memory and storage are both below 20\%. In particular, when being compared with MCUNet whose NN structures are optimized via NAS, AgileNN occupies the similar amount of memory but a much smaller amount of external storage. Such high memory and storage efficiency is particularly important on weak embedded devices with very limited storage resources, because it allows deployment of much more powerful NN models on these devices and hence provide solid support to more challenging NN applications. On the other hand, the SRAM usages of DeepCOD and SPINN are at the similar level to that of AgileNN.

\begin{figure*}
	\centering
	\hspace{0.4in}
	\subfigure[CIFAR-100 achieved skewness] { 
	\includegraphics[width=0.22\textwidth]{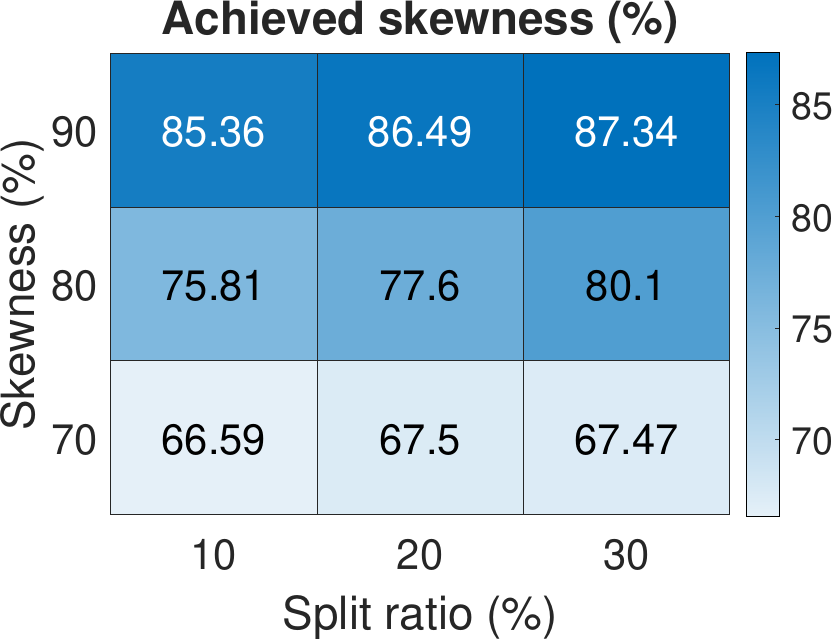}
	\label{fig:cifar100_skewness_sr_achieved_skewness}
}
	\hspace{0.3in}
	\subfigure[CIFAR-100 inference accuracy] { 
	\includegraphics[width=0.22\textwidth]{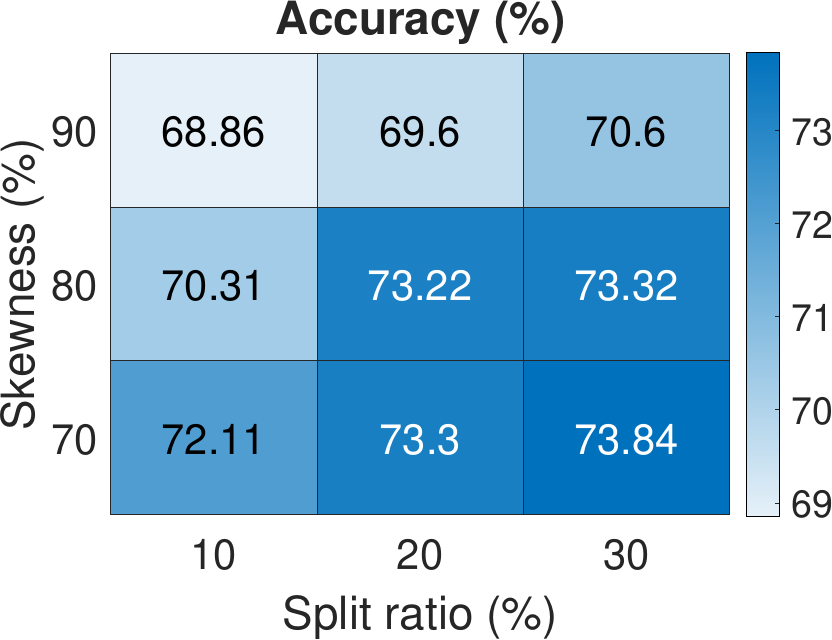}
	\label{fig:cifar100_skewness_sr_acc}
}
	\hspace{0.3in}
	\subfigure[CIFAR-100 network latency] { 
		\includegraphics[width=0.22\textwidth]{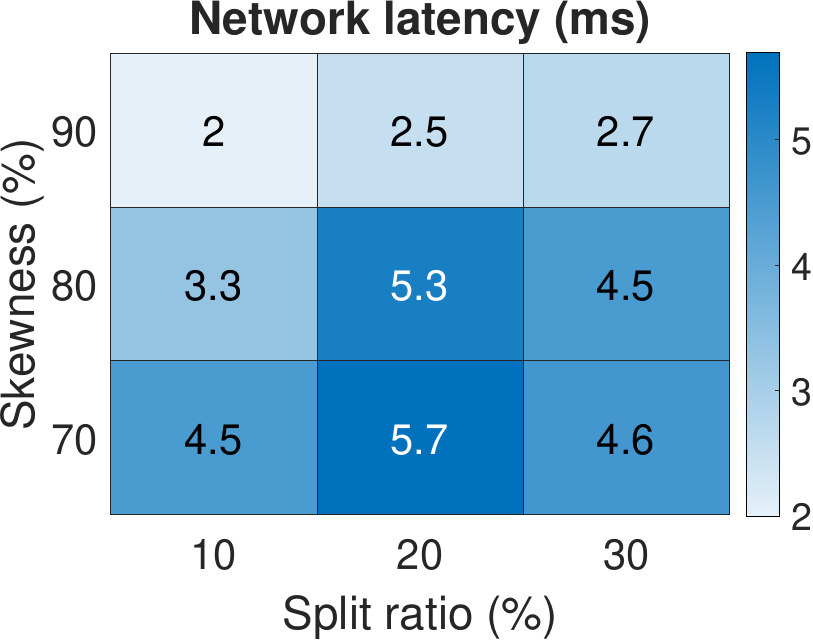}
		\label{fig:cifar100_skewness_sr_network_latency}
	}\newline
	
	\centering
		\subfigure[SVHN achieved skewness] { 
		\includegraphics[width=0.22\textwidth]{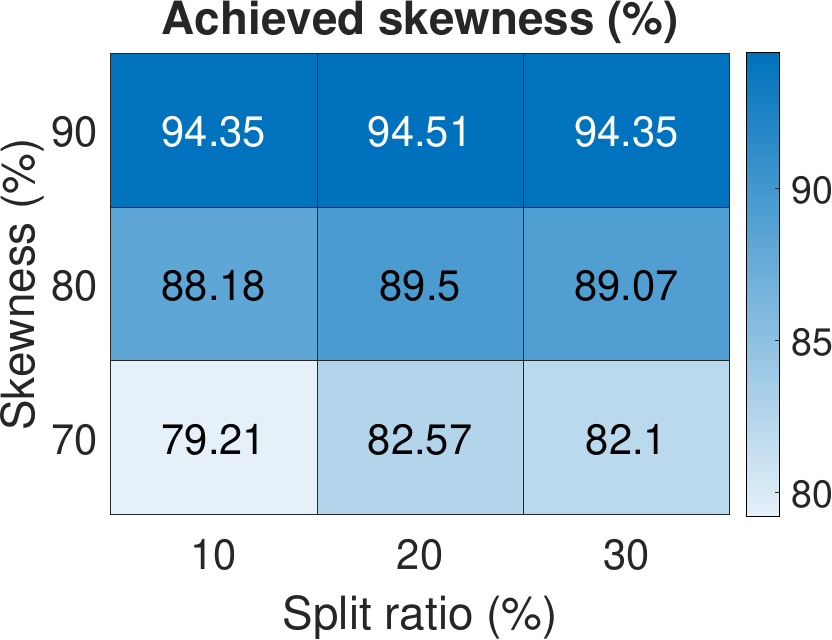}
		\label{fig:svhn_skewness_sr_achieved_skewness}
	}
			\hspace{0.3in}
\subfigure[SVHN inference accuracy] { 
	\includegraphics[width=0.22\textwidth]{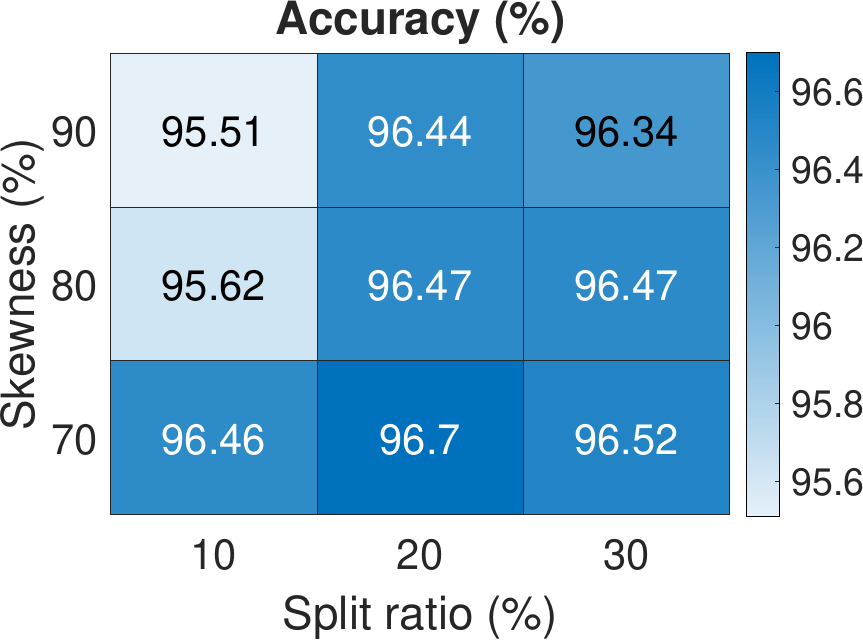}
	\label{fig:svhn_skewness_sr_acc}
}
	\hspace{0.3in}
	\subfigure[SVHN network latency] { 
		\includegraphics[width=0.22\textwidth]{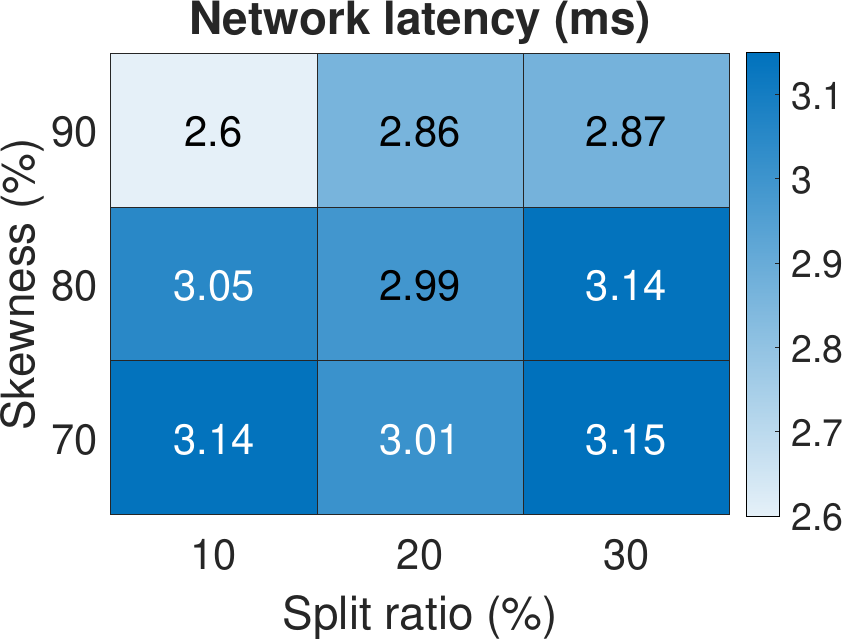}
		\label{fig:svhn_skewness_sr_network_latency}
	}
	\vspace{-0.15in}
	\caption{Effectiveness of skewness manipulation with different requirements of feature importance skewness}
	\label{fig:sr_skewness}
	\vspace{-0.1in}
\end{figure*}

\subsection{Effectiveness of Skewness Manipulation}
Skewness manipulation is the cornerstone of efficient NN offloading in AgileNN. To investigate the effectiveness of AgileNN's skewness manipulation, we apply different requirements of feature importance skewness by varying the value of $k$ between 3, 5 and 7, to retain 10\%, 20\% and 30\% of features with the highest importance at the local NN. Correspondingly, we require the the normalized importances of these features to reach 70\%, 80\% and 90\%, respectively.

We first verify whether AgileNN's skewness manipulation can adequately achieve the required skewness in the extracted features. Figure \ref{fig:cifar100_skewness_sr_achieved_skewness} and \ref{fig:svhn_skewness_sr_achieved_skewness} show that AgileNN can always meet the required skewness objective with minor difference. Especially on the SVHN dataset, the achieved skewness is even 4-12\% higher than the objective. This demonstrates that our skewness loss function described in Section 3.1 is highly effective.

Second, Figure\ref{fig:cifar100_skewness_sr_network_latency} and \ref{fig:svhn_skewness_sr_network_latency} show that, with the same amount of important features being retained at the local NN, enforcing higher skewness on these features can increase the feature sparsity on the remaining less important features, hence reducing the network transmission latency. At the same time, such higher skewness also affects the NN inference accuracy as shown in Figure \ref{fig:cifar100_skewness_sr_acc} and \ref{fig:svhn_skewness_sr_acc}. The major reason is that, when the normalized importances of locally retained features are too high, the lightweight Local NN may not have sufficient representation power to correctly perceive these features, hence leading to extra accuracy loss. However, since the Local and Remote NNs are jointly trained, such accuracy drop can be always constrained within 3\%. 

These results demonstrate that AgileNN can effectively manipulate the skewness of feature importance in different settings, hence allowing flexible tradeoffs between the accuracy and cost of NN inference. Retaining more features at the local devices could help mitigate such accuracy drop, at the expense of extra local NN computations. In practice, the optimal choice of skewness requirement and split ratio will depend on the specific device's computation power and characteristics of the training dataset. We generally suggest that the optimal design choice is to retain 20\% important features at the local device and require the normalized importance of these features to be $>$80\%. Such skewness requirement will be used in all the following experiments.



\begin{figure}[ht]
	\centering
	\vspace{-0.2in}
	\hspace{-0.25in}
	\subfigure[CIFAR-100] { 
		\includegraphics[width=0.22\textwidth]{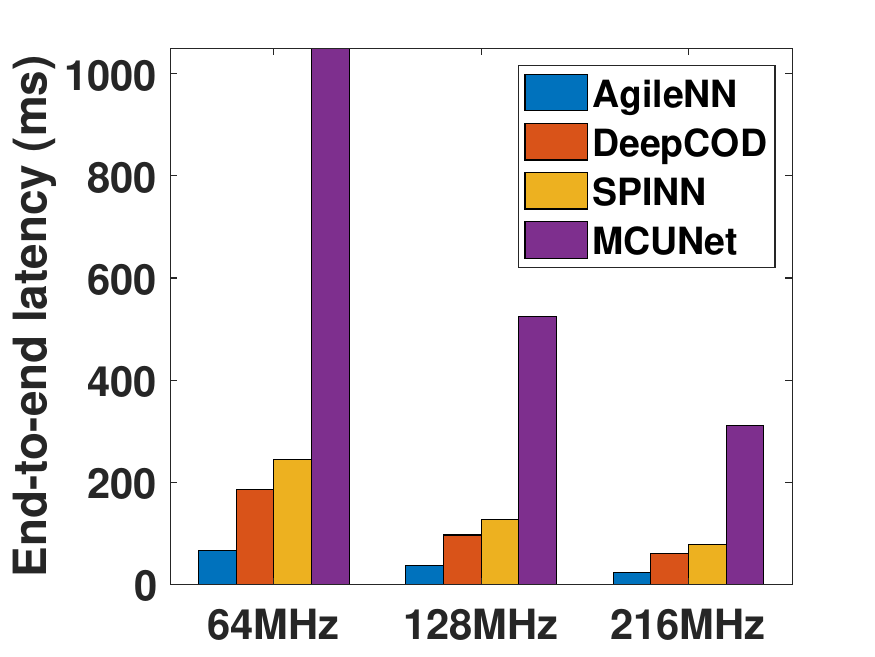}
		\label{fig:cifar100_freq_latency}
	}
	\subfigure[SVHN] { 
		\includegraphics[width=0.22\textwidth]{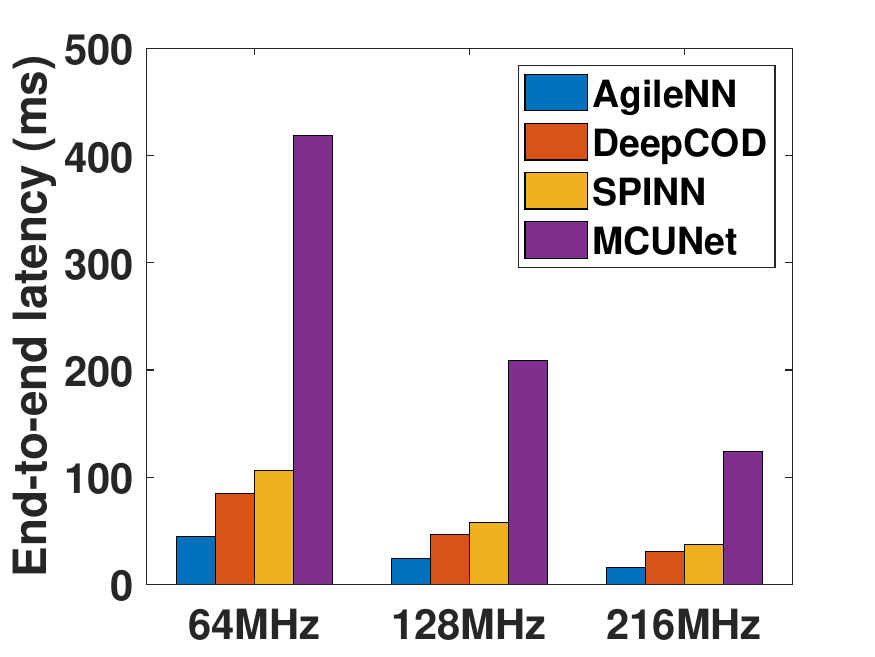}
		\label{fig:svhn_freq_latency}
	}
	\hspace{-0.4in}
	\vspace{-0.1in}
	\caption{The impact of different CPU Frequencies}
	\label{fig:cpu_freq}
	\vspace{-0.1in}
\end{figure}

\subsection{Impact of Local CPU Frequency}
Embedded devices may have CPUs with different frequencies. For example, the Arduino Nano uses an ATmega328 CPU at 16MHz and the STM32H743 MCU uses a dual-core ARM Cortex-M7 CPU at 480MHz, and the CPU frequency can also be adaptively configured at runtime. To study the impact of CPU frequency on AgileNN's performance, we adjust the CPU frequency of STM32F746 board by tuning its clock scaling factor. Here, we assume that most embedded devices, such as MCUs, will be exclusively used for NN inference when undertaking related computing tasks. Hence, we consider that the local device's CPU can be fully utilized for NN inference.

As shown in Figure \ref{fig:cpu_freq}, although the inference latency increases when the CPU frequency drops, such increase is always small when the CPU frequency drops from 216MHz to 64MHz, due to its lightweight feature exactor and Local NN. Comparatively, existing schemes suffer much higher performance degradation by running an expensive Local NN at the embedded device. For example, inference latency of MCUNet, SPINN and DeepCOD increased by 250\%, 200\% and 210\%, respectively.

\begin{figure}[ht]
	\centering
	\vspace{-0.15in}
	\hspace{-0.25in}
	\subfigure[CIFAR-100] { 
		\includegraphics[width=0.22\textwidth]{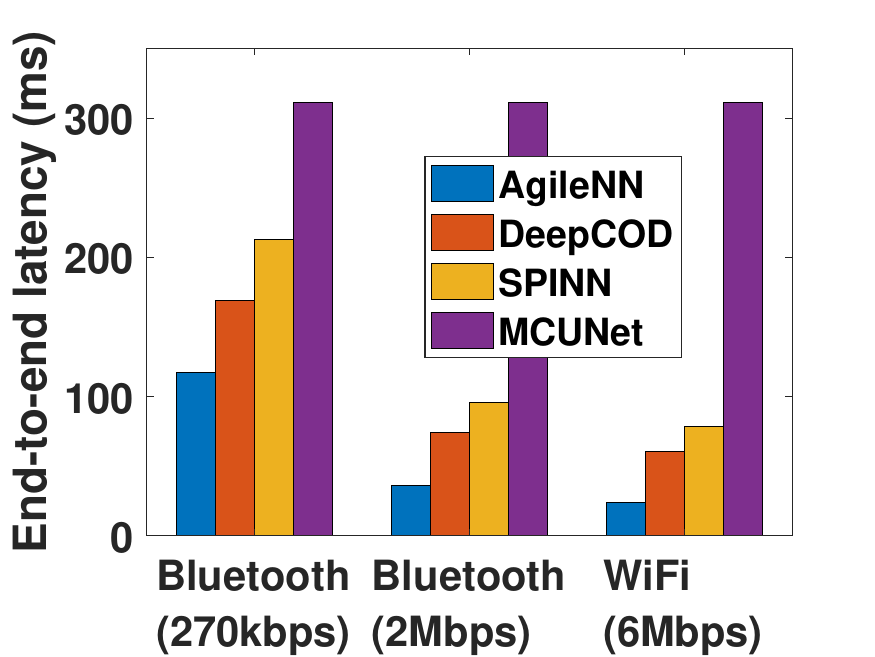}
		\label{fig:cifar100_bandwidth}
	}
	\subfigure[SVHN] { 
		\includegraphics[width=0.22\textwidth]{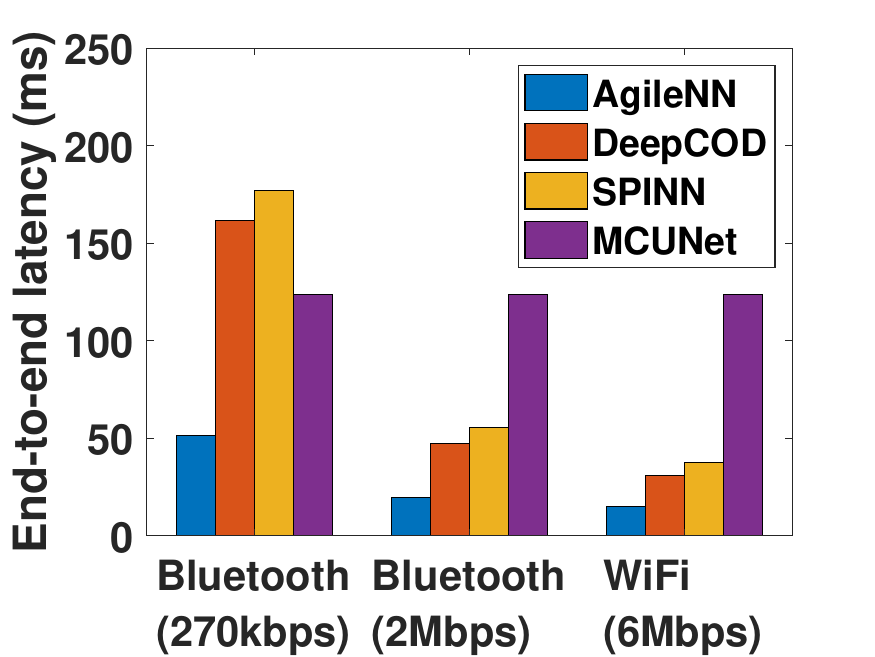}
		\label{fig:svhn_bandwidth}
	}
	\hspace{-0.4in}
	\vspace{-0.15in}
	\caption{The impact of different wireless bandwidths}
	\label{fig:bandwidth}
	\vspace{-0.1in}
\end{figure}

\subsection{Impact of Network Bandwidth}
Due to local constraints on power consumption and form factor, not all the embedded devices are equipped with high-speed WiFi modules. Instead, many of them have to use narrowband low-energy radios such as Bluetooth. Results in Figure \ref{fig:bandwidth} show that even when the available wireless network bandwidth is only 270kbps (95.5\% lower than that of WiFi), AgileNN's high feature sparsity ensures that it can still restrain the NN inference latency to be 50ms on the SVHN dataset and 100ms on the CIFAR-100 dataset. In contrast, the inference latency of DeepCOD and SPINN is largely dependent on the  wireless network bandwidth. These results imply that AgileNN outperforms other existing approaches in dynamic conditions of the wireless link connecting the local device and the server.

\begin{figure}[ht]
	\centering
	\hspace{-0.25in}
	\subfigure[Inference accuracy] { 
		\includegraphics[width=0.22\textwidth]{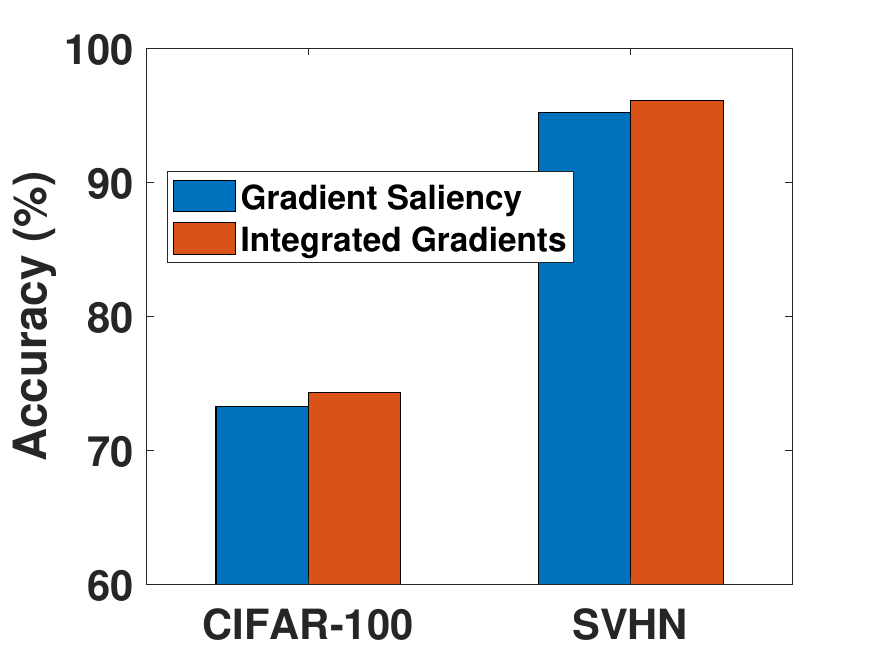}
		\label{fig:xai_choices_acc}
	}
	\subfigure[End-to-end latency] { 
		\includegraphics[width=0.22\textwidth]{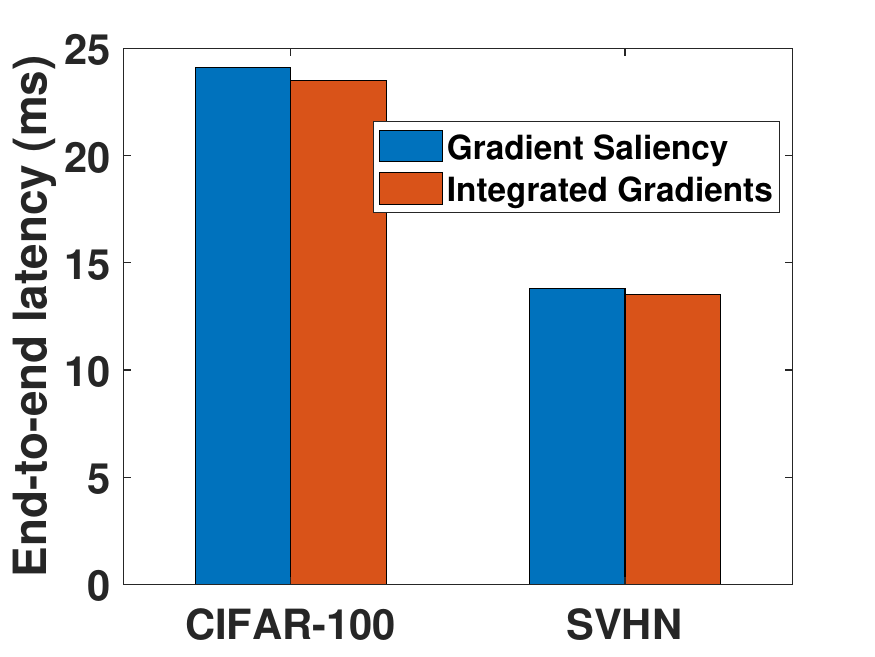}
		\label{fig:xai_choices_latency}
	}
	\hspace{-0.4in}
	\vspace{-0.15in}
	\caption{Different XAI techniques}
	\label{fig:xai_choices}
	\vspace{-0.1in}
\end{figure}

\subsection{Choices of XAI techniques}
The accuracy of importance evaluation varies with different XAI tools being used. To study such impact, we use two popular XAI tools: Gradient Saliency (GS) \cite{davies2021advancing} and Integrated Gradients (IG) \cite{sundararajan2017axiomatic} to construct AgileNN. As shown in Figure \ref{fig:xai_choices}, the performance of AgileNN remains stable with different XAI choices. IG makes AgileNN perform slightly better because it aggregates more interpolations of NN outputs' gradients as described in Section 2.2. On the other hand, IG is more computationally expensive because it usually requires 20-100 times of gradient computations to obtain each importance measurement. 

\section{Related Work}
\noindent\textbf{AI Attribution.} AgileNN leverages current NN attribution tools to evaluate feature importance. Traditional attribution approaches apply random permutation \cite{breiman2001random} or zero masks \cite{ribeiro2016should} to specific input variables, and use the induced output variation to empirically indicate importance. Attention-based approaches \cite{vaswani2017attention, zhao2020exploring} embed a learning-based weighting layer into the NN, and the learned weights are used to indicate feature importance. However, these measurements are sensitive to different NN structures and cannot always ensure accurate evaluation. 

Recent XAI techniques provide more accurate and robust attribution tools \cite{selvaraju2017grad, sundararajan2017axiomatic}. They adopt NN output's gradients with respect to the input variables to derive importance, which is more fine-grained and can clearly tell in percentage how much each input variable contributes to the output value. XAI techniques are mainly used for analyzing data characteristics and understanding NN behavior, but its usage for improving offloading efficiency is rarely explored by the existing work.

\noindent\textbf{On-device NN Inference.}
AgileNN is related to existing efforts on building lightweight NN models. NN compression \cite{denton2014exploiting, gong2014compressing} and pruning \cite{niu2020patdnn, han2015learning, fang2018nestdnn} tailor complicated NNs by removing redundant weights and structures. Neural Architecture Search (NAS) \cite{lin2020mcunet,banbury2021micronets} pushes it to the theoretical limit by searching for the optimal NN structure under the NN complexity constraint. In certain circumstances where wireless connectivity is unavailable at the local embedded device and local inference is hence the only option, these techniques could be useful to support some simple NN inference tasks with low performance requirements. However, due to the extreme resource constraints on weak embedded devices, these techniques have limited capability in supporting more complicated NN inferences or achieving real-time NN inference.

AgileNN is also related to recent work of NN offloading. Early efforts transmit the compressed raw data to the server \cite{ liu2018deepn,liu2019machine}. To improve data compressibility, later work adopts a local NN that transforms the raw data into sparse features \cite{eshratifar2019jointdnn, li2018jalad, yao2020deep, laskaridis2020spinn}, but the local NN should be complicated to ensure feature sparsity. Being orthogonal to AgileNN, there is work \cite{hsu2019couper, zhang2021elf} choosing to offload data to multiple servers to explore the heterogeneity of servers' computing power.

\section{Discussions}
\noindent\textbf{Reducing the training overhead.} Using XAI to evaluate the feature importance is computationally expensive, due to frequent computation of gradients in every training iteration. A straightforward mitigation is to reduce the amount of such gradient computations, but this may affect the quality of skewness manipulation. Alternatively, since standard NN training also involves gradient operations, it's possible to reuse these existing gradients to speed up XAI evaluation. We also expect the AI community to develop more lightweight XAI techniques in the near future. 



\noindent\textbf{Extreme network conditions.} As shown in Figure \ref{fig:bandwidth}, AgileNN outperforms the existing schemes when the available network bandwidth is low. If the network is unavailable or encounters strong interference, AgileNN can still rely on the local predictor to make basic decisions. Because the most important features are undertaken by the local predictor, AgileNN makes the best effort to maintain inference accuracy. It is also viable to deploy more complicated local predictors to improve accuracy under such extreme conditions.

\noindent\textbf{Other inference tasks.} In evaluations of this paper, we mainly target image recognition tasks, but AgileNN can also be applied to other inference tasks such as video and audio analytics. In particular, due to the limit memory capacity at weak embedded devices, it may be difficult to take the entire video as one NN input (e.g., video summarization) if the video size is large, but instead the video could be split and analyzed in segments. Each video segment, then, can be processed in a per-frame basis on the local device, and the video analytic task hence falls back to an image recognition task. Similarly, audio data can be converted into a 2D spectrum, which can be treated as images for NN inference.

\noindent\textbf{Offloading assisted training.} Although AgileNN speeds up the AI inference on weak devices, it is hard for static NN models to adopt to new data and different application scenarios. Instead, the NN model should be promptly retrained at run-time with the new incoming data, while incurring the minimum computation costs. AgileNN can be possibly extended to online training by incorporating a federated learning framework \cite{konevcny2016federated}, where multiple clients talk to a server without exposing local data. In this case, intermediate training results are forwarded to the server, which will then undertake majority of training overhead. Such extension of AgileNN will be our future work.

\section{Conclusion}
In this paper, we present AgileNN, a new technique that shifts the rationale of NN partitioning and offloading from fixed to agile and data-centric by leveraging the XAI techniques. AgileNN ensures real-time and accurate NN inference on extremely weak devices by migrating the required computations in NN offloading from online inference to offline training, and reduces the NN inference latency by up to 6$\times$ with similar accuracy compared to existing schemes.

\section*{Acknowledgments}
We thank the anonymous shepherd and reviewers for their comments and feedback. This work was supported in part by National Science Foundation (NSF) under grant number CNS-1812407, CNS-2029520, IIS-1956002, IIS-2205360, CCF-2217003 and CCF-2215042.

\bibliographystyle{ACM-Reference-Format}
\bibliography{ref}


\begin{thebibliography}{70}


\ifx \showCODEN    \undefined \def \showCODEN     #1{\unskip}     \fi
\ifx \showDOI      \undefined \def \showDOI       #1{#1}\fi
\ifx \showISBNx    \undefined \def \showISBNx     #1{\unskip}     \fi
\ifx \showISBNxiii \undefined \def \showISBNxiii  #1{\unskip}     \fi
\ifx \showISSN     \undefined \def \showISSN      #1{\unskip}     \fi
\ifx \showLCCN     \undefined \def \showLCCN      #1{\unskip}     \fi
\ifx \shownote     \undefined \def \shownote      #1{#1}          \fi
\ifx \showarticletitle \undefined \def \showarticletitle #1{#1}   \fi
\ifx \showURL      \undefined \def \showURL       {\relax}        \fi
\providecommand\bibfield[2]{#2}
\providecommand\bibinfo[2]{#2}
\providecommand\natexlab[1]{#1}
\providecommand\showeprint[2][]{arXiv:#2}

\bibitem[\protect\citeauthoryear{??}{pro}{[n. d.]}]%
        {progressiveautomations}
 \bibinfo{year}{[n. d.]}\natexlab{}.
\newblock \bibinfo{title}{Progressive Automations}.
\newblock
  \bibinfo{howpublished}{\url{https://www.progressiveautomations.com/pages/industrial-linear-actuators}}.
\newblock


\bibitem[\protect\citeauthoryear{??}{stm}{[n. d.]}]%
        {stm32f7}
 \bibinfo{year}{[n. d.]}\natexlab{}.
\newblock \bibinfo{title}{STM32F7/H7 Series Manual}.
\newblock
  \bibinfo{howpublished}{\url{https://www.st.com/resource/en/programming_manual/pm0253-stm32f7-series-and-stm32h7-series-cortexm7-processor-programming-manual-stmicroelectronics.pdf}}.
\newblock


\bibitem[\protect\citeauthoryear{Abdelhamid, Chen, Cho, Chandrakasan, and
  Adib}{Abdelhamid et~al\mbox{.}}{2020}]%
        {abdelhamid2020self}
\bibfield{author}{\bibinfo{person}{Mohamed~R Abdelhamid},
  \bibinfo{person}{Ruicong Chen}, \bibinfo{person}{Joonhyuk Cho},
  \bibinfo{person}{Anantha~P Chandrakasan}, {and} \bibinfo{person}{Fadel
  Adib}.} \bibinfo{year}{2020}\natexlab{}.
\newblock \showarticletitle{Self-reconfigurable micro-implants for cross-tissue
  wireless and batteryless connectivity}. In
  \bibinfo{booktitle}{\emph{MobiCom'20: Proceedings of the 26th Annual
  International Conference on Mobile Computing and Networking}}.
\newblock


\bibitem[\protect\citeauthoryear{Agustsson, Mentzer, Tschannen, Cavigelli,
  Timofte, Benini, and Gool}{Agustsson et~al\mbox{.}}{2017}]%
        {agustsson2017soft}
\bibfield{author}{\bibinfo{person}{Eirikur Agustsson}, \bibinfo{person}{Fabian
  Mentzer}, \bibinfo{person}{Michael Tschannen}, \bibinfo{person}{Lukas
  Cavigelli}, \bibinfo{person}{Radu Timofte}, \bibinfo{person}{Luca Benini},
  {and} \bibinfo{person}{Luc~V Gool}.} \bibinfo{year}{2017}\natexlab{}.
\newblock \showarticletitle{Soft-to-hard vector quantization for end-to-end
  learning compressible representations}.
\newblock \bibinfo{journal}{\emph{Advances in neural information processing
  systems}}  \bibinfo{volume}{30} (\bibinfo{year}{2017}).
\newblock


\bibitem[\protect\citeauthoryear{Ahmed, Jones, and Marks}{Ahmed
  et~al\mbox{.}}{2015}]%
        {ahmed2015improved}
\bibfield{author}{\bibinfo{person}{Ejaz Ahmed}, \bibinfo{person}{Michael
  Jones}, {and} \bibinfo{person}{Tim~K Marks}.}
  \bibinfo{year}{2015}\natexlab{}.
\newblock \showarticletitle{An improved deep learning architecture for person
  re-identification}. In \bibinfo{booktitle}{\emph{Proceedings of the IEEE
  conference on computer vision and pattern recognition}}.
  \bibinfo{pages}{3908--3916}.
\newblock


\bibitem[\protect\citeauthoryear{Amodei, Ananthanarayanan, Anubhai, Bai,
  Battenberg, Case, Casper, Catanzaro, Cheng, Chen, et~al\mbox{.}}{Amodei
  et~al\mbox{.}}{2016}]%
        {amodei2016deep}
\bibfield{author}{\bibinfo{person}{Dario Amodei}, \bibinfo{person}{Sundaram
  Ananthanarayanan}, \bibinfo{person}{Rishita Anubhai},
  \bibinfo{person}{Jingliang Bai}, \bibinfo{person}{Eric Battenberg},
  \bibinfo{person}{Carl Case}, \bibinfo{person}{Jared Casper},
  \bibinfo{person}{Bryan Catanzaro}, \bibinfo{person}{Qiang Cheng},
  \bibinfo{person}{Guoliang Chen}, {et~al\mbox{.}}}
  \bibinfo{year}{2016}\natexlab{}.
\newblock \showarticletitle{Deep speech 2: End-to-end speech recognition in
  english and mandarin}. In \bibinfo{booktitle}{\emph{International conference
  on machine learning}}. PMLR, \bibinfo{pages}{173--182}.
\newblock


\bibitem[\protect\citeauthoryear{Ayyalasomayajula, Arun, Wu, Sharma, Sethi,
  Vasisht, and Bharadia}{Ayyalasomayajula et~al\mbox{.}}{2020}]%
        {ayyalasomayajula2020deep}
\bibfield{author}{\bibinfo{person}{Roshan Ayyalasomayajula},
  \bibinfo{person}{Aditya Arun}, \bibinfo{person}{Chenfeng Wu},
  \bibinfo{person}{Sanatan Sharma}, \bibinfo{person}{Abhishek~Rajkumar Sethi},
  \bibinfo{person}{Deepak Vasisht}, {and} \bibinfo{person}{Dinesh Bharadia}.}
  \bibinfo{year}{2020}\natexlab{}.
\newblock \showarticletitle{Deep learning based wireless localization for
  indoor navigation}. In \bibinfo{booktitle}{\emph{Proceedings of the 26th
  Annual International Conference on Mobile Computing and Networking}}.
  \bibinfo{pages}{1--14}.
\newblock


\bibitem[\protect\citeauthoryear{Bahdanau, Cho, and Bengio}{Bahdanau
  et~al\mbox{.}}{2014}]%
        {bahdanau2014neural}
\bibfield{author}{\bibinfo{person}{Dzmitry Bahdanau},
  \bibinfo{person}{Kyunghyun Cho}, {and} \bibinfo{person}{Yoshua Bengio}.}
  \bibinfo{year}{2014}\natexlab{}.
\newblock \showarticletitle{Neural machine translation by jointly learning to
  align and translate}.
\newblock \bibinfo{journal}{\emph{arXiv preprint arXiv:1409.0473}}
  (\bibinfo{year}{2014}).
\newblock


\bibitem[\protect\citeauthoryear{Bakker and Huijsing}{Bakker and
  Huijsing}{1999}]%
        {bakker1999low}
\bibfield{author}{\bibinfo{person}{Anton Bakker} {and} \bibinfo{person}{Johan~H
  Huijsing}.} \bibinfo{year}{1999}\natexlab{}.
\newblock \showarticletitle{A low-cost high-accuracy CMOS smart temperature
  sensor}. In \bibinfo{booktitle}{\emph{Proceedings of the 25th European
  Solid-State Circuits Conference}}. IEEE, \bibinfo{pages}{302--305}.
\newblock


\bibitem[\protect\citeauthoryear{Banbury, Zhou, Fedorov, Matas, Thakker, Gope,
  Janapa~Reddi, Mattina, and Whatmough}{Banbury et~al\mbox{.}}{2021}]%
        {banbury2021micronets}
\bibfield{author}{\bibinfo{person}{Colby Banbury}, \bibinfo{person}{Chuteng
  Zhou}, \bibinfo{person}{Igor Fedorov}, \bibinfo{person}{Ramon Matas},
  \bibinfo{person}{Urmish Thakker}, \bibinfo{person}{Dibakar Gope},
  \bibinfo{person}{Vijay Janapa~Reddi}, \bibinfo{person}{Matthew Mattina},
  {and} \bibinfo{person}{Paul Whatmough}.} \bibinfo{year}{2021}\natexlab{}.
\newblock \showarticletitle{Micronets: Neural network architectures for
  deploying tinyml applications on commodity microcontrollers}.
\newblock \bibinfo{journal}{\emph{Proceedings of Machine Learning and Systems}}
   \bibinfo{volume}{3} (\bibinfo{year}{2021}).
\newblock


\bibitem[\protect\citeauthoryear{Bottou}{Bottou}{2012}]%
        {bottou2012stochastic}
\bibfield{author}{\bibinfo{person}{L{\'e}on Bottou}.}
  \bibinfo{year}{2012}\natexlab{}.
\newblock \showarticletitle{Stochastic gradient descent tricks}.
\newblock In \bibinfo{booktitle}{\emph{Neural networks: Tricks of the trade}}.
  \bibinfo{publisher}{Springer}, \bibinfo{pages}{421--436}.
\newblock


\bibitem[\protect\citeauthoryear{Breiman}{Breiman}{2001}]%
        {breiman2001random}
\bibfield{author}{\bibinfo{person}{Leo Breiman}.}
  \bibinfo{year}{2001}\natexlab{}.
\newblock \showarticletitle{Random forests}.
\newblock \bibinfo{journal}{\emph{Machine learning}} \bibinfo{volume}{45},
  \bibinfo{number}{1} (\bibinfo{year}{2001}), \bibinfo{pages}{5--32}.
\newblock


\bibitem[\protect\citeauthoryear{Bui, Pham, Barnitz, Zou, Nguyen, Truong, Kim,
  Farrow, Nguyen, Xiao, et~al\mbox{.}}{Bui et~al\mbox{.}}{2019}]%
        {bui2019ebp}
\bibfield{author}{\bibinfo{person}{Nam Bui}, \bibinfo{person}{Nhat Pham},
  \bibinfo{person}{Jessica~Jacqueline Barnitz}, \bibinfo{person}{Zhanan Zou},
  \bibinfo{person}{Phuc Nguyen}, \bibinfo{person}{Hoang Truong},
  \bibinfo{person}{Taeho Kim}, \bibinfo{person}{Nicholas Farrow},
  \bibinfo{person}{Anh Nguyen}, \bibinfo{person}{Jianliang Xiao},
  {et~al\mbox{.}}} \bibinfo{year}{2019}\natexlab{}.
\newblock \showarticletitle{ebp: A wearable system for frequent and comfortable
  blood pressure monitoring from user's ear}. In \bibinfo{booktitle}{\emph{The
  25th annual international conference on mobile computing and networking}}.
  \bibinfo{pages}{1--17}.
\newblock


\bibitem[\protect\citeauthoryear{Chen, Badrinarayanan, Lee, and
  Rabinovich}{Chen et~al\mbox{.}}{2018}]%
        {chen2018gradnorm}
\bibfield{author}{\bibinfo{person}{Zhao Chen}, \bibinfo{person}{Vijay
  Badrinarayanan}, \bibinfo{person}{Chen-Yu Lee}, {and} \bibinfo{person}{Andrew
  Rabinovich}.} \bibinfo{year}{2018}\natexlab{}.
\newblock \showarticletitle{Gradnorm: Gradient normalization for adaptive loss
  balancing in deep multitask networks}. In
  \bibinfo{booktitle}{\emph{International Conference on Machine Learning}}.
  PMLR, \bibinfo{pages}{794--803}.
\newblock


\bibitem[\protect\citeauthoryear{Ciaparrone, S{\'a}nchez, Tabik, Troiano,
  Tagliaferri, and Herrera}{Ciaparrone et~al\mbox{.}}{2020}]%
        {ciaparrone2020deep}
\bibfield{author}{\bibinfo{person}{Gioele Ciaparrone},
  \bibinfo{person}{Francisco~Luque S{\'a}nchez}, \bibinfo{person}{Siham Tabik},
  \bibinfo{person}{Luigi Troiano}, \bibinfo{person}{Roberto Tagliaferri}, {and}
  \bibinfo{person}{Francisco Herrera}.} \bibinfo{year}{2020}\natexlab{}.
\newblock \showarticletitle{Deep learning in video multi-object tracking: A
  survey}.
\newblock \bibinfo{journal}{\emph{Neurocomputing}}  \bibinfo{volume}{381}
  (\bibinfo{year}{2020}), \bibinfo{pages}{61--88}.
\newblock


\bibitem[\protect\citeauthoryear{Davies, Veli{\v{c}}kovi{\'c}, Buesing,
  Blackwell, Zheng, Toma{\v{s}}ev, Tanburn, Battaglia, Blundell, Juh{\'a}sz,
  et~al\mbox{.}}{Davies et~al\mbox{.}}{2021}]%
        {davies2021advancing}
\bibfield{author}{\bibinfo{person}{Alex Davies}, \bibinfo{person}{Petar
  Veli{\v{c}}kovi{\'c}}, \bibinfo{person}{Lars Buesing}, \bibinfo{person}{Sam
  Blackwell}, \bibinfo{person}{Daniel Zheng}, \bibinfo{person}{Nenad
  Toma{\v{s}}ev}, \bibinfo{person}{Richard Tanburn}, \bibinfo{person}{Peter
  Battaglia}, \bibinfo{person}{Charles Blundell}, \bibinfo{person}{Andr{\'a}s
  Juh{\'a}sz}, {et~al\mbox{.}}} \bibinfo{year}{2021}\natexlab{}.
\newblock \showarticletitle{Advancing mathematics by guiding human intuition
  with AI}.
\newblock \bibinfo{journal}{\emph{Nature}} \bibinfo{volume}{600},
  \bibinfo{number}{7887} (\bibinfo{year}{2021}), \bibinfo{pages}{70--74}.
\newblock


\bibitem[\protect\citeauthoryear{Deng, Dong, Socher, Li, Li, and Fei-Fei}{Deng
  et~al\mbox{.}}{2009}]%
        {deng2009imagenet}
\bibfield{author}{\bibinfo{person}{Jia Deng}, \bibinfo{person}{Wei Dong},
  \bibinfo{person}{Richard Socher}, \bibinfo{person}{Li-Jia Li},
  \bibinfo{person}{Kai Li}, {and} \bibinfo{person}{Li Fei-Fei}.}
  \bibinfo{year}{2009}\natexlab{}.
\newblock \showarticletitle{Imagenet: A large-scale hierarchical image
  database}. In \bibinfo{booktitle}{\emph{2009 IEEE conference on computer
  vision and pattern recognition}}. Ieee, \bibinfo{pages}{248--255}.
\newblock


\bibitem[\protect\citeauthoryear{Denton, Zaremba, Bruna, LeCun, and
  Fergus}{Denton et~al\mbox{.}}{2014}]%
        {denton2014exploiting}
\bibfield{author}{\bibinfo{person}{Emily~L Denton}, \bibinfo{person}{Wojciech
  Zaremba}, \bibinfo{person}{Joan Bruna}, \bibinfo{person}{Yann LeCun}, {and}
  \bibinfo{person}{Rob Fergus}.} \bibinfo{year}{2014}\natexlab{}.
\newblock \showarticletitle{Exploiting linear structure within convolutional
  networks for efficient evaluation}. In \bibinfo{booktitle}{\emph{Advances in
  neural information processing systems}}. \bibinfo{pages}{1269--1277}.
\newblock


\bibitem[\protect\citeauthoryear{Diaz, Heirich, Khider, and Robertson}{Diaz
  et~al\mbox{.}}{2013}]%
        {diaz2013optimal}
\bibfield{author}{\bibinfo{person}{Estefania~Munoz Diaz},
  \bibinfo{person}{Oliver Heirich}, \bibinfo{person}{Mohammed Khider}, {and}
  \bibinfo{person}{Patrick Robertson}.} \bibinfo{year}{2013}\natexlab{}.
\newblock \showarticletitle{Optimal sampling frequency and bias error modeling
  for foot-mounted IMUs}. In \bibinfo{booktitle}{\emph{International Conference
  on Indoor Positioning and Indoor Navigation}}. IEEE, \bibinfo{pages}{1--9}.
\newblock


\bibitem[\protect\citeauthoryear{Eshratifar, Abrishami, and Pedram}{Eshratifar
  et~al\mbox{.}}{2019}]%
        {eshratifar2019jointdnn}
\bibfield{author}{\bibinfo{person}{Amir~Erfan Eshratifar},
  \bibinfo{person}{Mohammad~Saeed Abrishami}, {and} \bibinfo{person}{Massoud
  Pedram}.} \bibinfo{year}{2019}\natexlab{}.
\newblock \showarticletitle{JointDNN: An efficient training and inference
  engine for intelligent mobile cloud computing services}.
\newblock \bibinfo{journal}{\emph{IEEE Transactions on Mobile Computing}}
  \bibinfo{volume}{20}, \bibinfo{number}{2} (\bibinfo{year}{2019}),
  \bibinfo{pages}{565--576}.
\newblock


\bibitem[\protect\citeauthoryear{Fang, Zeng, and Zhang}{Fang
  et~al\mbox{.}}{2018}]%
        {fang2018nestdnn}
\bibfield{author}{\bibinfo{person}{Biyi Fang}, \bibinfo{person}{Xiao Zeng},
  {and} \bibinfo{person}{Mi Zhang}.} \bibinfo{year}{2018}\natexlab{}.
\newblock \showarticletitle{Nestdnn: Resource-aware multi-tenant on-device deep
  learning for continuous mobile vision}. In
  \bibinfo{booktitle}{\emph{Proceedings of the 24th Annual International
  Conference on Mobile Computing and Networking}}. \bibinfo{pages}{115--127}.
\newblock


\bibitem[\protect\citeauthoryear{Frankle, Dziugaite, Roy, and Carbin}{Frankle
  et~al\mbox{.}}{2020}]%
        {frankle2020pruning}
\bibfield{author}{\bibinfo{person}{Jonathan Frankle},
  \bibinfo{person}{Gintare~Karolina Dziugaite}, \bibinfo{person}{Daniel~M Roy},
  {and} \bibinfo{person}{Michael Carbin}.} \bibinfo{year}{2020}\natexlab{}.
\newblock \showarticletitle{Pruning neural networks at initialization: Why are
  we missing the mark?}
\newblock \bibinfo{journal}{\emph{arXiv preprint arXiv:2009.08576}}
  (\bibinfo{year}{2020}).
\newblock


\bibitem[\protect\citeauthoryear{Gao, Hsu, Lee, Shen, and Subramanian}{Gao
  et~al\mbox{.}}{2017}]%
        {gao2017intention}
\bibfield{author}{\bibinfo{person}{Wei Gao}, \bibinfo{person}{David Hsu},
  \bibinfo{person}{Wee~Sun Lee}, \bibinfo{person}{Shengmei Shen}, {and}
  \bibinfo{person}{Karthikk Subramanian}.} \bibinfo{year}{2017}\natexlab{}.
\newblock \showarticletitle{Intention-net: Integrating planning and deep
  learning for goal-directed autonomous navigation}. In
  \bibinfo{booktitle}{\emph{Conference on robot learning}}. PMLR,
  \bibinfo{pages}{185--194}.
\newblock


\bibitem[\protect\citeauthoryear{Gobieski, Lucia, and Beckmann}{Gobieski
  et~al\mbox{.}}{2019}]%
        {gobieski2019intelligence}
\bibfield{author}{\bibinfo{person}{Graham Gobieski}, \bibinfo{person}{Brandon
  Lucia}, {and} \bibinfo{person}{Nathan Beckmann}.}
  \bibinfo{year}{2019}\natexlab{}.
\newblock \showarticletitle{Intelligence beyond the edge: Inference on
  intermittent embedded systems}. In \bibinfo{booktitle}{\emph{Proceedings of
  the Twenty-Fourth International Conference on Architectural Support for
  Programming Languages and Operating Systems}}. \bibinfo{pages}{199--213}.
\newblock


\bibitem[\protect\citeauthoryear{Gong, Liu, Yang, and Bourdev}{Gong
  et~al\mbox{.}}{2014}]%
        {gong2014compressing}
\bibfield{author}{\bibinfo{person}{Yunchao Gong}, \bibinfo{person}{Liu Liu},
  \bibinfo{person}{Ming Yang}, {and} \bibinfo{person}{Lubomir Bourdev}.}
  \bibinfo{year}{2014}\natexlab{}.
\newblock \showarticletitle{Compressing deep convolutional networks using
  vector quantization}.
\newblock \bibinfo{journal}{\emph{arXiv preprint arXiv:1412.6115}}
  (\bibinfo{year}{2014}).
\newblock


\bibitem[\protect\citeauthoryear{Groenendijk, Karaoglu, Gevers, and
  Mensink}{Groenendijk et~al\mbox{.}}{2021}]%
        {groenendijk2021multi}
\bibfield{author}{\bibinfo{person}{Rick Groenendijk}, \bibinfo{person}{Sezer
  Karaoglu}, \bibinfo{person}{Theo Gevers}, {and} \bibinfo{person}{Thomas
  Mensink}.} \bibinfo{year}{2021}\natexlab{}.
\newblock \showarticletitle{Multi-loss weighting with coefficient of
  variations}. In \bibinfo{booktitle}{\emph{Proceedings of the IEEE/CVF Winter
  Conference on Applications of Computer Vision}}. \bibinfo{pages}{1469--1478}.
\newblock


\bibitem[\protect\citeauthoryear{Han, Pool, Tran, and Dally}{Han
  et~al\mbox{.}}{2015}]%
        {han2015learning}
\bibfield{author}{\bibinfo{person}{Song Han}, \bibinfo{person}{Jeff Pool},
  \bibinfo{person}{John Tran}, {and} \bibinfo{person}{William~J Dally}.}
  \bibinfo{year}{2015}\natexlab{}.
\newblock \showarticletitle{Learning both weights and connections for efficient
  neural networks}.
\newblock \bibinfo{journal}{\emph{arXiv preprint arXiv:1506.02626}}
  (\bibinfo{year}{2015}).
\newblock


\bibitem[\protect\citeauthoryear{He, Zhang, Ren, and Sun}{He
  et~al\mbox{.}}{2016}]%
        {he2016deep}
\bibfield{author}{\bibinfo{person}{Kaiming He}, \bibinfo{person}{Xiangyu
  Zhang}, \bibinfo{person}{Shaoqing Ren}, {and} \bibinfo{person}{Jian Sun}.}
  \bibinfo{year}{2016}\natexlab{}.
\newblock \showarticletitle{Deep residual learning for image recognition}. In
  \bibinfo{booktitle}{\emph{Proceedings of the IEEE conference on computer
  vision and pattern recognition}}. \bibinfo{pages}{770--778}.
\newblock


\bibitem[\protect\citeauthoryear{Hesse, Schaub-Meyer, and Roth}{Hesse
  et~al\mbox{.}}{2021}]%
        {hesse2021fast}
\bibfield{author}{\bibinfo{person}{Robin Hesse}, \bibinfo{person}{Simone
  Schaub-Meyer}, {and} \bibinfo{person}{Stefan Roth}.}
  \bibinfo{year}{2021}\natexlab{}.
\newblock \showarticletitle{Fast axiomatic attribution for neural networks}.
\newblock \bibinfo{journal}{\emph{Advances in Neural Information Processing
  Systems}}  \bibinfo{volume}{34} (\bibinfo{year}{2021}),
  \bibinfo{pages}{19513--19524}.
\newblock


\bibitem[\protect\citeauthoryear{Hsu, Bhardwaj, and Gavrilovska}{Hsu
  et~al\mbox{.}}{2019}]%
        {hsu2019couper}
\bibfield{author}{\bibinfo{person}{Ke-Jou Hsu}, \bibinfo{person}{Ketan
  Bhardwaj}, {and} \bibinfo{person}{Ada Gavrilovska}.}
  \bibinfo{year}{2019}\natexlab{}.
\newblock \showarticletitle{Couper: Dnn model slicing for visual analytics
  containers at the edge}. In \bibinfo{booktitle}{\emph{Proceedings of the 4th
  ACM/IEEE Symposium on Edge Computing}}. \bibinfo{pages}{179--194}.
\newblock


\bibitem[\protect\citeauthoryear{Hu, Bao, Wang, and Liu}{Hu
  et~al\mbox{.}}{2019}]%
        {hu2019dynamic}
\bibfield{author}{\bibinfo{person}{Chuang Hu}, \bibinfo{person}{Wei Bao},
  \bibinfo{person}{Dan Wang}, {and} \bibinfo{person}{Fengming Liu}.}
  \bibinfo{year}{2019}\natexlab{}.
\newblock \showarticletitle{Dynamic adaptive DNN surgery for inference
  acceleration on the edge}. In \bibinfo{booktitle}{\emph{IEEE INFOCOM
  2019-IEEE Conference on Computer Communications}}. IEEE,
  \bibinfo{pages}{1423--1431}.
\newblock


\bibitem[\protect\citeauthoryear{Hu, Yang, Yi, Kittler, Christmas, Li, and
  Hospedales}{Hu et~al\mbox{.}}{2015}]%
        {hu2015face}
\bibfield{author}{\bibinfo{person}{Guosheng Hu}, \bibinfo{person}{Yongxin
  Yang}, \bibinfo{person}{Dong Yi}, \bibinfo{person}{Josef Kittler},
  \bibinfo{person}{William Christmas}, \bibinfo{person}{Stan~Z Li}, {and}
  \bibinfo{person}{Timothy Hospedales}.} \bibinfo{year}{2015}\natexlab{}.
\newblock \showarticletitle{When face recognition meets with deep learning: an
  evaluation of convolutional neural networks for face recognition}. In
  \bibinfo{booktitle}{\emph{Proceedings of the IEEE international conference on
  computer vision workshops}}. \bibinfo{pages}{142--150}.
\newblock


\bibitem[\protect\citeauthoryear{Iyer, Talla, Kellogg, Gollakota, and
  Smith}{Iyer et~al\mbox{.}}{2016}]%
        {iyer2016inter}
\bibfield{author}{\bibinfo{person}{Vikram Iyer}, \bibinfo{person}{Vamsi Talla},
  \bibinfo{person}{Bryce Kellogg}, \bibinfo{person}{Shyamnath Gollakota}, {and}
  \bibinfo{person}{Joshua Smith}.} \bibinfo{year}{2016}\natexlab{}.
\newblock \showarticletitle{Inter-technology backscatter: Towards internet
  connectivity for implanted devices}. In \bibinfo{booktitle}{\emph{Proceedings
  of the 2016 ACM SIGCOMM Conference}}. \bibinfo{pages}{356--369}.
\newblock


\bibitem[\protect\citeauthoryear{Kang, Hauswald, Gao, Rovinski, Mudge, Mars,
  and Tang}{Kang et~al\mbox{.}}{2017}]%
        {kang2017neurosurgeon}
\bibfield{author}{\bibinfo{person}{Yiping Kang}, \bibinfo{person}{Johann
  Hauswald}, \bibinfo{person}{Cao Gao}, \bibinfo{person}{Austin Rovinski},
  \bibinfo{person}{Trevor Mudge}, \bibinfo{person}{Jason Mars}, {and}
  \bibinfo{person}{Lingjia Tang}.} \bibinfo{year}{2017}\natexlab{}.
\newblock \showarticletitle{Neurosurgeon: Collaborative intelligence between
  the cloud and mobile edge}.
\newblock \bibinfo{journal}{\emph{ACM SIGARCH Computer Architecture News}}
  \bibinfo{volume}{45}, \bibinfo{number}{1} (\bibinfo{year}{2017}),
  \bibinfo{pages}{615--629}.
\newblock


\bibitem[\protect\citeauthoryear{Kellogg, Parks, Gollakota, Smith, and
  Wetherall}{Kellogg et~al\mbox{.}}{2014}]%
        {kellogg2014wi}
\bibfield{author}{\bibinfo{person}{Bryce Kellogg}, \bibinfo{person}{Aaron
  Parks}, \bibinfo{person}{Shyamnath Gollakota}, \bibinfo{person}{Joshua~R
  Smith}, {and} \bibinfo{person}{David Wetherall}.}
  \bibinfo{year}{2014}\natexlab{}.
\newblock \showarticletitle{Wi-Fi backscatter: Internet connectivity for
  RF-powered devices}. In \bibinfo{booktitle}{\emph{Proceedings of the 2014 ACM
  Conference on SIGCOMM}}. \bibinfo{pages}{607--618}.
\newblock


\bibitem[\protect\citeauthoryear{Ko, Na, Amir, and Mukhopadhyay}{Ko
  et~al\mbox{.}}{2018}]%
        {ko2018edge}
\bibfield{author}{\bibinfo{person}{Jong~Hwan Ko}, \bibinfo{person}{Taesik Na},
  \bibinfo{person}{Mohammad~Faisal Amir}, {and} \bibinfo{person}{Saibal
  Mukhopadhyay}.} \bibinfo{year}{2018}\natexlab{}.
\newblock \showarticletitle{Edge-host partitioning of deep neural networks with
  feature space encoding for resource-constrained internet-of-things
  platforms}. In \bibinfo{booktitle}{\emph{2018 15th IEEE International
  Conference on Advanced Video and Signal Based Surveillance (AVSS)}}. IEEE,
  \bibinfo{pages}{1--6}.
\newblock


\bibitem[\protect\citeauthoryear{Kone{\v{c}}n{\`y}, McMahan, Yu, Richt{\'a}rik,
  Suresh, and Bacon}{Kone{\v{c}}n{\`y} et~al\mbox{.}}{2016}]%
        {konevcny2016federated}
\bibfield{author}{\bibinfo{person}{Jakub Kone{\v{c}}n{\`y}},
  \bibinfo{person}{H~Brendan McMahan}, \bibinfo{person}{Felix~X Yu},
  \bibinfo{person}{Peter Richt{\'a}rik}, \bibinfo{person}{Ananda~Theertha
  Suresh}, {and} \bibinfo{person}{Dave Bacon}.}
  \bibinfo{year}{2016}\natexlab{}.
\newblock \showarticletitle{Federated learning: Strategies for improving
  communication efficiency}.
\newblock \bibinfo{journal}{\emph{arXiv preprint arXiv:1610.05492}}
  (\bibinfo{year}{2016}).
\newblock


\bibitem[\protect\citeauthoryear{Krizhevsky, Hinton, et~al\mbox{.}}{Krizhevsky
  et~al\mbox{.}}{2009}]%
        {krizhevsky2009learning}
\bibfield{author}{\bibinfo{person}{Alex Krizhevsky}, \bibinfo{person}{Geoffrey
  Hinton}, {et~al\mbox{.}}} \bibinfo{year}{2009}\natexlab{}.
\newblock \showarticletitle{Learning multiple layers of features from tiny
  images}.
\newblock  (\bibinfo{year}{2009}).
\newblock


\bibitem[\protect\citeauthoryear{Laskaridis, Venieris, Almeida, Leontiadis, and
  Lane}{Laskaridis et~al\mbox{.}}{2020}]%
        {laskaridis2020spinn}
\bibfield{author}{\bibinfo{person}{Stefanos Laskaridis},
  \bibinfo{person}{Stylianos~I Venieris}, \bibinfo{person}{Mario Almeida},
  \bibinfo{person}{Ilias Leontiadis}, {and} \bibinfo{person}{Nicholas~D Lane}.}
  \bibinfo{year}{2020}\natexlab{}.
\newblock \showarticletitle{SPINN: synergistic progressive inference of neural
  networks over device and cloud}. In \bibinfo{booktitle}{\emph{Proceedings of
  the 26th Annual International Conference on Mobile Computing and
  Networking}}. \bibinfo{pages}{1--15}.
\newblock


\bibitem[\protect\citeauthoryear{Le and Yang}{Le and Yang}{2015}]%
        {le2015tiny}
\bibfield{author}{\bibinfo{person}{Ya Le} {and} \bibinfo{person}{Xuan Yang}.}
  \bibinfo{year}{2015}\natexlab{}.
\newblock \showarticletitle{Tiny imagenet visual recognition challenge}.
\newblock \bibinfo{journal}{\emph{CS 231N}} \bibinfo{volume}{7},
  \bibinfo{number}{7} (\bibinfo{year}{2015}), \bibinfo{pages}{3}.
\newblock


\bibitem[\protect\citeauthoryear{Le~Gall}{Le~Gall}{1991}]%
        {le1991mpeg}
\bibfield{author}{\bibinfo{person}{Didier Le~Gall}.}
  \bibinfo{year}{1991}\natexlab{}.
\newblock \showarticletitle{MPEG: A video compression standard for multimedia
  applications}.
\newblock \bibinfo{journal}{\emph{Commun. ACM}} \bibinfo{volume}{34},
  \bibinfo{number}{4} (\bibinfo{year}{1991}), \bibinfo{pages}{46--58}.
\newblock


\bibitem[\protect\citeauthoryear{Li, Hu, Jiang, Wang, Wen, and Zhu}{Li
  et~al\mbox{.}}{2018}]%
        {li2018jalad}
\bibfield{author}{\bibinfo{person}{Hongshan Li}, \bibinfo{person}{Chenghao Hu},
  \bibinfo{person}{Jingyan Jiang}, \bibinfo{person}{Zhi Wang},
  \bibinfo{person}{Yonggang Wen}, {and} \bibinfo{person}{Wenwu Zhu}.}
  \bibinfo{year}{2018}\natexlab{}.
\newblock \showarticletitle{Jalad: Joint accuracy-and latency-aware deep
  structure decoupling for edge-cloud execution}. In
  \bibinfo{booktitle}{\emph{2018 IEEE 24th international conference on parallel
  and distributed systems (ICPADS)}}. IEEE, \bibinfo{pages}{671--678}.
\newblock


\bibitem[\protect\citeauthoryear{Li and Zhang}{Li and Zhang}{2011}]%
        {li2011reducing}
\bibfield{author}{\bibinfo{person}{Yiran Li} {and} \bibinfo{person}{Tong
  Zhang}.} \bibinfo{year}{2011}\natexlab{}.
\newblock \showarticletitle{Reducing dram image data access energy consumption
  in video processing}.
\newblock \bibinfo{journal}{\emph{IEEE Transactions on Multimedia}}
  \bibinfo{volume}{14}, \bibinfo{number}{2} (\bibinfo{year}{2011}),
  \bibinfo{pages}{303--313}.
\newblock


\bibitem[\protect\citeauthoryear{Lin, Chen, Lin, Cohn, Gan, and Han}{Lin
  et~al\mbox{.}}{2020}]%
        {lin2020mcunet}
\bibfield{author}{\bibinfo{person}{Ji Lin}, \bibinfo{person}{Wei-Ming Chen},
  \bibinfo{person}{Yujun Lin}, \bibinfo{person}{John Cohn},
  \bibinfo{person}{Chuang Gan}, {and} \bibinfo{person}{Song Han}.}
  \bibinfo{year}{2020}\natexlab{}.
\newblock \showarticletitle{Mcunet: Tiny deep learning on iot devices}.
\newblock \bibinfo{journal}{\emph{arXiv preprint arXiv:2007.10319}}
  (\bibinfo{year}{2020}).
\newblock


\bibitem[\protect\citeauthoryear{Liu, Liu, Wen, Jiang, Xu, Wang, and Quan}{Liu
  et~al\mbox{.}}{2018}]%
        {liu2018deepn}
\bibfield{author}{\bibinfo{person}{Zihao Liu}, \bibinfo{person}{Tao Liu},
  \bibinfo{person}{Wujie Wen}, \bibinfo{person}{Lei Jiang},
  \bibinfo{person}{Jie Xu}, \bibinfo{person}{Yanzhi Wang}, {and}
  \bibinfo{person}{Gang Quan}.} \bibinfo{year}{2018}\natexlab{}.
\newblock \showarticletitle{DeepN-JPEG: A deep neural network favorable
  JPEG-based image compression framework}. In
  \bibinfo{booktitle}{\emph{Proceedings of the 55th annual design automation
  conference}}. \bibinfo{pages}{1--6}.
\newblock


\bibitem[\protect\citeauthoryear{Liu, Xu, Liu, Liu, Wang, Shi, Wen, Huang,
  Yuan, and Zhuang}{Liu et~al\mbox{.}}{2019}]%
        {liu2019machine}
\bibfield{author}{\bibinfo{person}{Zihao Liu}, \bibinfo{person}{Xiaowei Xu},
  \bibinfo{person}{Tao Liu}, \bibinfo{person}{Qi Liu}, \bibinfo{person}{Yanzhi
  Wang}, \bibinfo{person}{Yiyu Shi}, \bibinfo{person}{Wujie Wen},
  \bibinfo{person}{Meiping Huang}, \bibinfo{person}{Haiyun Yuan}, {and}
  \bibinfo{person}{Jian Zhuang}.} \bibinfo{year}{2019}\natexlab{}.
\newblock \showarticletitle{Machine vision guided 3d medical image compression
  for efficient transmission and accurate segmentation in the clouds}. In
  \bibinfo{booktitle}{\emph{Proceedings of the IEEE/CVF Conference on Computer
  Vision and Pattern Recognition}}. \bibinfo{pages}{12687--12696}.
\newblock


\bibitem[\protect\citeauthoryear{Ma, Luo, Steiger, Traverso, and Adib}{Ma
  et~al\mbox{.}}{2018}]%
        {ma2018enabling}
\bibfield{author}{\bibinfo{person}{Yunfei Ma}, \bibinfo{person}{Zhihong Luo},
  \bibinfo{person}{Christoph Steiger}, \bibinfo{person}{Giovanni Traverso},
  {and} \bibinfo{person}{Fadel Adib}.} \bibinfo{year}{2018}\natexlab{}.
\newblock \showarticletitle{Enabling deep-tissue networking for miniature
  medical devices}. In \bibinfo{booktitle}{\emph{Proceedings of the 2018
  Conference of the ACM Special Interest Group on Data Communication}}.
  \bibinfo{pages}{417--431}.
\newblock


\bibitem[\protect\citeauthoryear{Nair and Hinton}{Nair and Hinton}{2010}]%
        {nair2010rectified}
\bibfield{author}{\bibinfo{person}{Vinod Nair} {and}
  \bibinfo{person}{Geoffrey~E Hinton}.} \bibinfo{year}{2010}\natexlab{}.
\newblock \showarticletitle{Rectified linear units improve restricted boltzmann
  machines}. In \bibinfo{booktitle}{\emph{Proceedings of the 27th International
  Conference on Machine Learning (ICML)}}.
\newblock


\bibitem[\protect\citeauthoryear{Nelson}{Nelson}{1989}]%
        {nelson1989lzw}
\bibfield{author}{\bibinfo{person}{Mark~R Nelson}.}
  \bibinfo{year}{1989}\natexlab{}.
\newblock \showarticletitle{LZW data compression}.
\newblock \bibinfo{journal}{\emph{Dr. Dobb's Journal}} \bibinfo{volume}{14},
  \bibinfo{number}{10} (\bibinfo{year}{1989}), \bibinfo{pages}{29--36}.
\newblock


\bibitem[\protect\citeauthoryear{Netzer, Wang, Coates, Bissacco, Wu, and
  Ng}{Netzer et~al\mbox{.}}{2011}]%
        {netzer2011reading}
\bibfield{author}{\bibinfo{person}{Yuval Netzer}, \bibinfo{person}{Tao Wang},
  \bibinfo{person}{Adam Coates}, \bibinfo{person}{Alessandro Bissacco},
  \bibinfo{person}{Bo Wu}, {and} \bibinfo{person}{Andrew~Y Ng}.}
  \bibinfo{year}{2011}\natexlab{}.
\newblock \showarticletitle{Reading digits in natural images with unsupervised
  feature learning}.
\newblock  (\bibinfo{year}{2011}).
\newblock


\bibitem[\protect\citeauthoryear{Niu, Ma, Lin, Wang, Qian, Lin, Wang, and
  Ren}{Niu et~al\mbox{.}}{2020}]%
        {niu2020patdnn}
\bibfield{author}{\bibinfo{person}{Wei Niu}, \bibinfo{person}{Xiaolong Ma},
  \bibinfo{person}{Sheng Lin}, \bibinfo{person}{Shihao Wang},
  \bibinfo{person}{Xuehai Qian}, \bibinfo{person}{Xue Lin},
  \bibinfo{person}{Yanzhi Wang}, {and} \bibinfo{person}{Bin Ren}.}
  \bibinfo{year}{2020}\natexlab{}.
\newblock \showarticletitle{Patdnn: Achieving real-time dnn execution on mobile
  devices with pattern-based weight pruning}. In
  \bibinfo{booktitle}{\emph{Proceedings of the Twenty-Fifth International
  Conference on Architectural Support for Programming Languages and Operating
  Systems}}. \bibinfo{pages}{907--922}.
\newblock


\bibitem[\protect\citeauthoryear{Niu, Ma, Wang, and Ren}{Niu
  et~al\mbox{.}}{2019}]%
        {niu201926ms}
\bibfield{author}{\bibinfo{person}{Wei Niu}, \bibinfo{person}{Xiaolong Ma},
  \bibinfo{person}{Yanzhi Wang}, {and} \bibinfo{person}{Bin Ren}.}
  \bibinfo{year}{2019}\natexlab{}.
\newblock \showarticletitle{26ms inference time for resnet-50: Towards
  real-time execution of all dnns on smartphone}.
\newblock \bibinfo{journal}{\emph{arXiv preprint arXiv:1905.00571}}
  (\bibinfo{year}{2019}).
\newblock


\bibitem[\protect\citeauthoryear{Ribeiro, Singh, and Guestrin}{Ribeiro
  et~al\mbox{.}}{2016a}]%
        {ribeiro2016should}
\bibfield{author}{\bibinfo{person}{Marco~Tulio Ribeiro},
  \bibinfo{person}{Sameer Singh}, {and} \bibinfo{person}{Carlos Guestrin}.}
  \bibinfo{year}{2016}\natexlab{a}.
\newblock \showarticletitle{" Why should i trust you?" Explaining the
  predictions of any classifier}. In \bibinfo{booktitle}{\emph{Proceedings of
  the 22nd ACM SIGKDD international conference on knowledge discovery and data
  mining}}. \bibinfo{pages}{1135--1144}.
\newblock


\bibitem[\protect\citeauthoryear{Ribeiro, Singh, and Guestrin}{Ribeiro
  et~al\mbox{.}}{2016b}]%
        {ribeiro2016model}
\bibfield{author}{\bibinfo{person}{Marco~Tulio Ribeiro},
  \bibinfo{person}{Sameer Singh}, {and} \bibinfo{person}{Carlos Guestrin}.}
  \bibinfo{year}{2016}\natexlab{b}.
\newblock \showarticletitle{Model-agnostic interpretability of machine
  learning}.
\newblock \bibinfo{journal}{\emph{arXiv preprint arXiv:1606.05386}}
  (\bibinfo{year}{2016}).
\newblock


\bibitem[\protect\citeauthoryear{Sandler, Howard, Zhu, Zhmoginov, and
  Chen}{Sandler et~al\mbox{.}}{2018}]%
        {sandler2018mobilenetv2}
\bibfield{author}{\bibinfo{person}{Mark Sandler}, \bibinfo{person}{Andrew
  Howard}, \bibinfo{person}{Menglong Zhu}, \bibinfo{person}{Andrey Zhmoginov},
  {and} \bibinfo{person}{Liang-Chieh Chen}.} \bibinfo{year}{2018}\natexlab{}.
\newblock \showarticletitle{Mobilenetv2: Inverted residuals and linear
  bottlenecks}. In \bibinfo{booktitle}{\emph{Proceedings of the IEEE conference
  on computer vision and pattern recognition}}. \bibinfo{pages}{4510--4520}.
\newblock


\bibitem[\protect\citeauthoryear{Selvaraju, Cogswell, Das, Vedantam, Parikh,
  and Batra}{Selvaraju et~al\mbox{.}}{2017}]%
        {selvaraju2017grad}
\bibfield{author}{\bibinfo{person}{Ramprasaath~R Selvaraju},
  \bibinfo{person}{Michael Cogswell}, \bibinfo{person}{Abhishek Das},
  \bibinfo{person}{Ramakrishna Vedantam}, \bibinfo{person}{Devi Parikh}, {and}
  \bibinfo{person}{Dhruv Batra}.} \bibinfo{year}{2017}\natexlab{}.
\newblock \showarticletitle{Grad-cam: Visual explanations from deep networks
  via gradient-based localization}. In \bibinfo{booktitle}{\emph{Proceedings of
  the IEEE international conference on computer vision}}.
  \bibinfo{pages}{618--626}.
\newblock


\bibitem[\protect\citeauthoryear{Shrikumar, Greenside, and Kundaje}{Shrikumar
  et~al\mbox{.}}{2017}]%
        {ShrikumarGK17}
\bibfield{author}{\bibinfo{person}{Avanti Shrikumar}, \bibinfo{person}{Peyton
  Greenside}, {and} \bibinfo{person}{Anshul Kundaje}.}
  \bibinfo{year}{2017}\natexlab{}.
\newblock \showarticletitle{Learning Important Features Through Propagating
  Activation Differences}.
\newblock \bibinfo{journal}{\emph{CoRR}}  \bibinfo{volume}{abs/1704.02685}
  (\bibinfo{year}{2017}).
\newblock


\bibitem[\protect\citeauthoryear{Simonyan and Zisserman}{Simonyan and
  Zisserman}{2014}]%
        {simonyan2014very}
\bibfield{author}{\bibinfo{person}{Karen Simonyan} {and}
  \bibinfo{person}{Andrew Zisserman}.} \bibinfo{year}{2014}\natexlab{}.
\newblock \showarticletitle{Very deep convolutional networks for large-scale
  image recognition}.
\newblock \bibinfo{journal}{\emph{arXiv preprint arXiv:1409.1556}}
  (\bibinfo{year}{2014}).
\newblock


\bibitem[\protect\citeauthoryear{Sundararajan, Taly, and Yan}{Sundararajan
  et~al\mbox{.}}{2017}]%
        {sundararajan2017axiomatic}
\bibfield{author}{\bibinfo{person}{Mukund Sundararajan}, \bibinfo{person}{Ankur
  Taly}, {and} \bibinfo{person}{Qiqi Yan}.} \bibinfo{year}{2017}\natexlab{}.
\newblock \showarticletitle{Axiomatic attribution for deep networks}. In
  \bibinfo{booktitle}{\emph{International Conference on Machine Learning}}.
  PMLR, \bibinfo{pages}{3319--3328}.
\newblock


\bibitem[\protect\citeauthoryear{Tan and Le}{Tan and Le}{2021}]%
        {tan2021efficientnetv2}
\bibfield{author}{\bibinfo{person}{Mingxing Tan} {and} \bibinfo{person}{Quoc
  Le}.} \bibinfo{year}{2021}\natexlab{}.
\newblock \showarticletitle{Efficientnetv2: Smaller models and faster
  training}. In \bibinfo{booktitle}{\emph{International Conference on Machine
  Learning}}. PMLR, \bibinfo{pages}{10096--10106}.
\newblock


\bibitem[\protect\citeauthoryear{Vaswani, Shazeer, Parmar, Uszkoreit, Jones,
  Gomez, Kaiser, and Polosukhin}{Vaswani et~al\mbox{.}}{2017}]%
        {vaswani2017attention}
\bibfield{author}{\bibinfo{person}{Ashish Vaswani}, \bibinfo{person}{Noam
  Shazeer}, \bibinfo{person}{Niki Parmar}, \bibinfo{person}{Jakob Uszkoreit},
  \bibinfo{person}{Llion Jones}, \bibinfo{person}{Aidan~N Gomez},
  \bibinfo{person}{{\L}ukasz Kaiser}, {and} \bibinfo{person}{Illia
  Polosukhin}.} \bibinfo{year}{2017}\natexlab{}.
\newblock \showarticletitle{Attention is all you need}.
\newblock \bibinfo{journal}{\emph{Advances in neural information processing
  systems}}  \bibinfo{volume}{30} (\bibinfo{year}{2017}).
\newblock


\bibitem[\protect\citeauthoryear{Wallace}{Wallace}{1992}]%
        {wallace1992jpeg}
\bibfield{author}{\bibinfo{person}{Gregory~K Wallace}.}
  \bibinfo{year}{1992}\natexlab{}.
\newblock \showarticletitle{The JPEG still picture compression standard}.
\newblock \bibinfo{journal}{\emph{IEEE transactions on consumer electronics}}
  \bibinfo{volume}{38}, \bibinfo{number}{1} (\bibinfo{year}{1992}),
  \bibinfo{pages}{xviii--xxxiv}.
\newblock


\bibitem[\protect\citeauthoryear{Wang, Chen, Song, Guizani, Yu, and Du}{Wang
  et~al\mbox{.}}{2018}]%
        {wang2018iot}
\bibfield{author}{\bibinfo{person}{Dan Wang}, \bibinfo{person}{Dong Chen},
  \bibinfo{person}{Bin Song}, \bibinfo{person}{Nadra Guizani},
  \bibinfo{person}{Xiaoyan Yu}, {and} \bibinfo{person}{Xiaojiang Du}.}
  \bibinfo{year}{2018}\natexlab{}.
\newblock \showarticletitle{From IoT to 5G I-IoT: The next generation IoT-based
  intelligent algorithms and 5G technologies}.
\newblock \bibinfo{journal}{\emph{IEEE Communications Magazine}}
  \bibinfo{volume}{56}, \bibinfo{number}{10} (\bibinfo{year}{2018}),
  \bibinfo{pages}{114--120}.
\newblock


\bibitem[\protect\citeauthoryear{Wu, Wang, Jiang, Ye, and Xue}{Wu
  et~al\mbox{.}}{2015}]%
        {wu2015modeling}
\bibfield{author}{\bibinfo{person}{Zuxuan Wu}, \bibinfo{person}{Xi Wang},
  \bibinfo{person}{Yu-Gang Jiang}, \bibinfo{person}{Hao Ye}, {and}
  \bibinfo{person}{Xiangyang Xue}.} \bibinfo{year}{2015}\natexlab{}.
\newblock \showarticletitle{Modeling spatial-temporal clues in a hybrid deep
  learning framework for video classification}. In
  \bibinfo{booktitle}{\emph{Proceedings of the 23rd ACM international
  conference on Multimedia}}. \bibinfo{pages}{461--470}.
\newblock


\bibitem[\protect\citeauthoryear{Yao, Li, Liu, Wang, Liu, Shao, and
  Abdelzaher}{Yao et~al\mbox{.}}{2020}]%
        {yao2020deep}
\bibfield{author}{\bibinfo{person}{Shuochao Yao}, \bibinfo{person}{Jinyang Li},
  \bibinfo{person}{Dongxin Liu}, \bibinfo{person}{Tianshi Wang},
  \bibinfo{person}{Shengzhong Liu}, \bibinfo{person}{Huajie Shao}, {and}
  \bibinfo{person}{Tarek Abdelzaher}.} \bibinfo{year}{2020}\natexlab{}.
\newblock \showarticletitle{Deep compressive offloading: Speeding up neural
  network inference by trading edge computation for network latency}. In
  \bibinfo{booktitle}{\emph{Proceedings of the 18th Conference on Embedded
  Networked Sensor Systems}}. \bibinfo{pages}{476--488}.
\newblock


\bibitem[\protect\citeauthoryear{Yu, Yu, Cui, Tao, and Tian}{Yu
  et~al\mbox{.}}{2019}]%
        {yu2019deep}
\bibfield{author}{\bibinfo{person}{Zhou Yu}, \bibinfo{person}{Jun Yu},
  \bibinfo{person}{Yuhao Cui}, \bibinfo{person}{Dacheng Tao}, {and}
  \bibinfo{person}{Qi Tian}.} \bibinfo{year}{2019}\natexlab{}.
\newblock \showarticletitle{Deep modular co-attention networks for visual
  question answering}. In \bibinfo{booktitle}{\emph{Proceedings of the IEEE/CVF
  conference on computer vision and pattern recognition}}.
  \bibinfo{pages}{6281--6290}.
\newblock


\bibitem[\protect\citeauthoryear{Zhang, He, Liu, Jia, Liu, Gruteser,
  Raychaudhuri, and Zhang}{Zhang et~al\mbox{.}}{2021}]%
        {zhang2021elf}
\bibfield{author}{\bibinfo{person}{Wuyang Zhang}, \bibinfo{person}{Zhezhi He},
  \bibinfo{person}{Luyang Liu}, \bibinfo{person}{Zhenhua Jia},
  \bibinfo{person}{Yunxin Liu}, \bibinfo{person}{Marco Gruteser},
  \bibinfo{person}{Dipankar Raychaudhuri}, {and} \bibinfo{person}{Yanyong
  Zhang}.} \bibinfo{year}{2021}\natexlab{}.
\newblock \showarticletitle{Elf: accelerate high-resolution mobile deep vision
  with content-aware parallel offloading}. In
  \bibinfo{booktitle}{\emph{Proceedings of the 27th Annual International
  Conference on Mobile Computing and Networking}}. \bibinfo{pages}{201--214}.
\newblock


\bibitem[\protect\citeauthoryear{Zhao, Jia, and Koltun}{Zhao
  et~al\mbox{.}}{2020}]%
        {zhao2020exploring}
\bibfield{author}{\bibinfo{person}{Hengshuang Zhao}, \bibinfo{person}{Jiaya
  Jia}, {and} \bibinfo{person}{Vladlen Koltun}.}
  \bibinfo{year}{2020}\natexlab{}.
\newblock \showarticletitle{Exploring self-attention for image recognition}. In
  \bibinfo{booktitle}{\emph{Proceedings of the IEEE/CVF Conference on Computer
  Vision and Pattern Recognition}}. \bibinfo{pages}{10076--10085}.
\newblock


\bibitem[\protect\citeauthoryear{Zhao, Jiang, and Qiu}{Zhao
  et~al\mbox{.}}{2021}]%
        {zhao2021deep}
\bibfield{author}{\bibinfo{person}{Wentao Zhao}, \bibinfo{person}{Wei Jiang},
  {and} \bibinfo{person}{Xinguo Qiu}.} \bibinfo{year}{2021}\natexlab{}.
\newblock \showarticletitle{Deep learning for COVID-19 detection based on CT
  images}.
\newblock \bibinfo{journal}{\emph{Scientific Reports}} \bibinfo{volume}{11},
  \bibinfo{number}{1} (\bibinfo{year}{2021}), \bibinfo{pages}{1--12}.
\newblock


\bibitem[\protect\citeauthoryear{Zhao and Ye}{Zhao and Ye}{2008}]%
        {zhao2008low}
\bibfield{author}{\bibinfo{person}{Yanbo Zhao} {and} \bibinfo{person}{Zhaohui
  Ye}.} \bibinfo{year}{2008}\natexlab{}.
\newblock \showarticletitle{A low cost GSM/GPRS based wireless home security
  system}.
\newblock \bibinfo{journal}{\emph{IEEE Transactions on Consumer Electronics}}
  \bibinfo{volume}{54}, \bibinfo{number}{2} (\bibinfo{year}{2008}),
  \bibinfo{pages}{567--572}.
\newblock


\end{thebibliography}

\end{document}